\documentclass[authoryear,5p,times]{elsarticle}

\journal{Medical Image Analysis}
\pdfoutput=1

\usepackage{amsmath,amsfonts}
\usepackage{graphicx}
\usepackage{epstopdf}
\usepackage{float}
\usepackage{array,multirow}
\usepackage{algorithm}
\usepackage{soul}
\usepackage[colorlinks=true, allcolors=blue]{hyperref}

\newcommand{\z}{\mathbf{z}}
\newcommand{\h}{{\mathbf{H}^z}}
\newcommand{\y}{\mathbf{y}}
\newcommand{\vis}{\mathbf{v}}
\newcommand{\hid}{{\mathbf{H}^v}}
\newcommand{\g}{{\mathbf{H}^y}}
\newcommand{\data}{\mathbf{D}}
\newcommand{\labels}{{\mathbf{l}}}
\newcommand{\nodes}{{\boldsymbol \eta}}
\newcommand{\param}{{\boldsymbol \theta}}

\begin{document}

\begin{frontmatter}
\title{A Modality-Adaptive Method for Segmenting Brain Tumors and Organs-at-Risk in Radiation Therapy Planning} 

\author[a]{Mikael Agn \corref{mycorrespondingauthor}}
\cortext[mycorrespondingauthor]{Corresponding author}
\ead{miag@dtu.dk}
\author[i]{Per Munck af Rosensch\"{o}ld}
\author[c]{Oula Puonti}
\author[b]{Michael J. Lundemann}
\author[d,e]{Laura Mancini}
\author[d,e]{Anastasia Papadaki}
\author[d,e]{Steffi~Thust}
\author[f]{John Ashburner}
\author[g]{Ian Law}
\author[a,h]{Koen Van Leemput}

\address[a]{Department of Applied Mathematics and Computer Science, Technical University of Denmark, Denmark}
\address[i]{Radiation Physics, Department of Hematology, Oncology and Radiation Physics, Sk\r{a}ne University Hospital, Lund, Sweden}
\address[c]{Danish Research Centre for Magnetic Resonance, Copenhagen University Hospital Hvidovre, Denmark}
\address[b]{Department of Oncology, Copenhagen University Hospital Rigshospitalet, Denmark}
\address[d]{Neuroradiological Academic Unit, Department of Brain Repair and Rehabilitation, UCL Institute of Neurology, University College London, UK}
\address[e]{Lysholm Department of Neuroradiology, National Hospital for Neurology and Neurosurgery, UCLH NHS Foundation Trust, UK}
\address[f]{Wellcome Centre for Human Neuroimaging, UCL Institute of Neurology, University College London, UK}
\address[g]{Department of Clinical Physiology, Nuclear Medicine and PET, Copenhagen University Hospital Rigshospitalet, Denmark}
\address[h]{Athinoula A. Martinos Center for Biomedical Imaging, Massachusetts General Hospital, Harvard Medical School, USA}

\begin{abstract}
In this paper we present a method for simultaneously segmenting brain tumors and an extensive set of organs-at-risk for radiation therapy planning of glioblastomas. The method combines a contrast-adaptive generative model for whole-brain segmentation with a new spatial regularization model of tumor shape using convolutional restricted Boltzmann machines. We demonstrate experimentally that the method is able to adapt to image acquisitions that differ substantially from any available training data, ensuring its applicability across treatment sites; that its tumor segmentation accuracy is comparable to that of the current state of the art; and that it captures most organs-at-risk sufficiently well for radiation therapy planning purposes. The proposed method may be a valuable step towards automating the delineation of brain tumors and organs-at-risk in glioblastoma patients undergoing radiation therapy.
\end{abstract}


\end{frontmatter}

\section{Introduction}
Glioblastomas, which are the most common type of malignant tumors originating within the brain \citep{Bleeker2012}, are commonly treated with a combination of surgical resection, chemo-therapy and radiation therapy.
During radiation therapy, the patient is 
subjected to radiation beams, typically from different directions and with different intensity profiles, with the aim of maximizing the delivered radiation dose to the targeted tumor while minimizing the dose to sensitive healthy structures, so-called organs-at-risk (OARs) \citep{Shaffer2010}. 
For the purpose of planning a radiation therapy session, these structures 
need to be delineated on 
computed tomography (CT) or magnetic resonance (MR) scans of the patient's head~\citep{munck2011}.

In current clinical practice, delineation is performed manually with limited assistance from automatic procedures, which is time consuming for the human expert and typically suffers from high inter-rater variability \citep{deelay2011, dolz2015c, menze2015}. 
These limitations are 
amplified in
emerging techniques for image-guided radiation therapy, which introduce a demand for continuous delineation during treatment \citep{lagendijk2014}. 
Consequently,
there is 
an increasing 
need for fast automated segmentation methods that can 
robustly segment both brain tumors and OARs
from clinically acquired head scans.

Recent years have seen an influx of \textit{discriminative} methods for brain tumor segmentation, with good -- although not very robust -- performance
reported in the annual MICCAI Brain Tumor Segmentation (BRATS) challenges~\citep{menze2015}. 
Discriminative
methods directly exploit the intensity information of annotated training data 
to discern between 
tumorous and 
other tissue in new images. 
Traditionally, they rely on user-engineered image features that are then fed into classifiers, such as random forests \citep{zikic2012, islam2013, tustison2015, meier2016} or support vector machines \citep{bauer2011}. 
Lately, however, convolutional neural networks (CNNs), 
which learn suitable image features  
simultaneously with their classifiers,
have become more prominent
\citep{pereira2016, kamnitsas2016,havaei2016}. 

Although discriminative methods have demonstrated state-of-the-art tumor segmentation performance, they suffer from 
several
drawbacks that limit their practical applicability in 
radiation therapy planning
settings. 
In particular,
what is needed 
in radiation therapy
is an accurate segmentation not just of the tumor, but also of a multitude of OARs.
Although CNNs 
segmenting dozens of brain substructures 
have recently been demonstrated 
\citep{roy2017error, rajchl2018neuronet}, 
using such methods 
in the context of radiation therapy planning 
is complicated by their need for large annotated training datasets,
as scans with
high-quality 
segmentations of 
both tumors and OARs in 
hundreds
of patients are not easily available.
Further exacerbating this issue is that
both
the 
type and the number of acquired images 
often
differ 
substantially
among treatment centers, 
not only 
as a result of 
differences in 
imaging
protocols 
and scanner platforms,
but also 
because of
the continuous development
of novel MR pulse 
sequences
for brain tumor imaging \citep{mabray2015modern, Sauwen2016}.
Although an active research area in the field \citep{havaei2016hemis,ghafoorian2017transfer,valindria2018domain}, 
effectively dealing with the ensuing
explosion of possible contrasts and contrast combinations remains an open problem
for discriminative segmentation methods.

In order to sidestep these difficulties with discriminative approaches, we present a method in this paper for simultaneously segmenting brain tumors and OARs using a \emph{generative} approach, in which prior knowledge of anatomy and the imaging process are 
incorporated
using Bayesian statistics.
Specifically,
our
method combines an existing contrast-adaptive method for whole-brain segmentation \citep{puonti2016} with a new spatial regularization model of tumor shape using generative neural networks. 
The OARs we consider in this paper are eyes, optic chiasm, optic nerves, brainstem, and hippocampi, but more structures can easily be added.
Compared to existing work, the proposed method presents several novel contributions:
\smallskip
\begin{enumerate}
  \item 
  To the best of our knowledge, this is the first  
  method that addresses the segmentation of both brain tumors and OARs within the same modeling framework.
  While existing generative methods for tumor segmentation typically 
  also perform
  classification into white matter, gray matter and cerebrospinal fluid \citep{moon2002automatic, prastawa2003automatic, menze2010, gooya2012, kwon2014, bakas2016},
  they do not 
  further 
  subdivide these tissue types into OARs, nor do they segment 
  OARs
  outside the brain.
  Conversely, with the exception of
  the optic system \citep{bekes2008, Noble2011, dolz2015b}, 
  most automated segmentation methods for OARs in radiation therapy applications have been concentrated on label transfer using non-linear registration of manually annotated template data \citep{dawant1999brain, cuadra2004, isambert2008, bauer2013, bondiau2005atlas}, 
  which does not address the problem of tumor segmentation itself.
  \vspace{-1mm}
  \item \phantom{xxx} 
  By adopting a generative approach, the proposed method makes judicious use of readily available training data. In particular, 
  the approach
  allows merging of disparate models of normal head structures, learned from manually annotated scans of normal subjects, with models of tumor shape derived from brain tumor patients,
  without requiring that segmentations of these two set of structures are available within the same set of subjects. 
  Importantly,
  once trained the same method can 
  be readily applied to data 
  from 
  different 
  sites
  without 
  retraining. 
  As we will demonstrate, this is the case even when
  data acquisitions 
  are fundamentally different from the data used to train the method, 
  such as
  CT scans 
  or
  experimental MR contrasts.
  \vspace{-5mm}
  \item \phantom{xxx}
  In contrast to discriminative methods for brain lesion segmentation, in which large spatial contexts are exploited to achieve state-of-the-art segmentation accuracy \citep{corso2008efficient, geremia2011spatial, karimaghaloo2016adaptive, brosch2016deep, kamnitsas2016}, 
  spatial 
  regularization of lesions in generative methods has so far been limited to local properties, such as local lesion probability in lesion-seeded probabilistic atlases \citep{moon2002automatic, prastawa2003automatic, gooya2012, kwon2014, bakas2016} or first-order Markov random fields (MRFs) in which only pairwise interactions between neighboring voxels are taken into account \citep{van2001automated, menze2010}.
  In this paper, we explore the potential of convolutional restricted Boltzmann machines (cRBMs) 
  \citep{lee2009}
  to provide long-range spatial regularization through MRFs with high-order clique potentials that are automatically learned from manual segmentations of brain tumors.
  We empirically demonstrate that these higher-order shape models yield an advantage in segmentation accuracy compared to first-order MRFs. 
\end{enumerate}
\smallskip
\noindent
Preliminary versions of 
this work appeared in  
two conference contributions \citep{agn2016b,Agn2016}. 
Here we 
extend 
the method 
to handle 
more OARs, in particular optic nerves, optic chiasm, and eyes;
describe the model and the statistical inference in more detail;
and 
provide an in-depth validation on a large number of patients, 
evaluating the method's adaptability to varying input data and suitability for radiation therapy planning.  

\section{Modeling framework}
\label{sec:modelNew}
\begin{figure}
\begin{center}
\includegraphics[width=0.7\columnwidth]{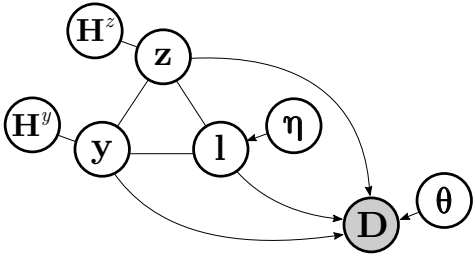}
\caption{Graphical representation of the model. The atlas-based prior on $\mathbf{l}$ is defined by parameters $\boldsymbol \eta$ governing the deformation of the atlas. The tumor-affected map $\mathbf{z}$ and the tumor core map $\mathbf{y}$ are connected to auxiliary variables $\mathbf{H}^z$ and $\mathbf{H}^y$, respectively. The variables $\mathbf{l}$, $\mathbf{z}$ and $\mathbf{y}$ jointly predict the data $\mathbf{D}$ according to the likelihood parameters $\boldsymbol \theta$.
Shading indicates observed variables.}
\label{f:fullmodel}
\end{center}
\end{figure}

Let $\mathbf{D} = (\mathbf{d}_1,...,\mathbf{d}_I)$ denote the data of $N$ co-registered medical images of a patient's head, where $I$ is the number of image voxels and $\mathbf{d}_i$ contains the log-transformed\footnote{We work with log-transformed intensities to model the MR bias field effect as an additive (rather than multiplicative) process, see Section \ref{s:likeh}.} intensities at voxel $i$. 
Each voxel $i$ has a normal label $l_i \in \{1,...,K\}$ that is associated with one of $K=17$ normal head structures, detailed in Table \ref{t:lab}, where $B$ denotes a set of structures located inside the brain. 
A voxel $i$ can be tumor-affected, indicated by $z_i=1$, where $z_i \in \{0,1\}$. Within tumor-affected tissue, a voxel $i$ can be either \textit{edema} or \textit{core}, indicated by $y_i=0$ and $y_i=1$, respectively, where $y_i \in \{0,1\}$. Edema corresponds to the visible peritumoral edema surrounding the core, which corresponds to the gross tumor volume (GTV) used in radiation therapy. To model the labels $l_i$, $z_i$ and $y_i$ across all voxels, we build a generative model that describes the image formation process, seen in Figure \ref{f:fullmodel}. The model consists of two parts.
The first part is a likelihood function $ p(\data|\labels,\z,\y, \param)$ 
that links the labels to image intensities, where $\labels = (l_1, ..., l_I)$, $\z = (z_1, ..., z_I)$, and $\y = (y_1, ..., y_I)$. This likelihood function depends on a set of parameters $\param$, governed by a prior distribution $p(\param)$, that allows
the model to adapt to images with different contrast properties. 
The second part is a segmentation prior $p(\labels,\z,\y | \nodes) = \sum_\h \sum_\g p(\labels,\z,\y,\h,\g | \nodes)$, where $\nodes$, with prior $p (\nodes)$, are parameters governing the deformation of a probabilistic atlas, and $\h$ and $\g$ are auxiliary variables that help encode high-order shape models of $\z$ and $\y$. 
\begin{table}
\caption{Labels associated with normal head structures, with brain structures in $B$.}
\label{t:lab}
\center
\footnotesize
\begin{tabular}{l l}
\hline
\hline
\vspace*{-2mm} \\
$l \in B$ & \{white matter (WM), grey matter (GM), cerebrospinal fluid \\
             &   (CSF), brainstem, unspecified brain tissue, and left and right\\
             &   hippocampus\} \\
\vspace*{-2mm} \\
$l \notin B$ & \{background, eye socket fat, eye socket muscles, optic chiasm; \\
                & and left and right optic nerve, eye tissue and eye fluid\}\\
\hline
\end{tabular}
\end{table}

We use this model to obtain a fully automated segmentation algorithm by evaluating the posterior of the labels given the data: 
\begin{equation}
p(\labels,\z,\y | \data) \propto p(\data|\labels,\z,\y) p(\labels,\z,\y),
\label{e:segmpost}
\end{equation}
where 
$
p(\labels,\z,\y) = \int_\nodes p(\labels,\z,\y | \nodes) p(\nodes) \mathrm{d} \nodes
$
and 
$
p(\data | \labels,\z,\y) = $ \linebreak $\int_\param p(\data|\labels,\z,\y, \param) p(\param) \mathrm{d} \param  
$ 
will be detailed in Section \ref{s:prior} and Section \ref{s:likeh}, respectively,
and computationally evaluating Eq.~\ref{e:segmpost} will be addressed in Section \ref{s:infer}.

\subsection{Segmentation prior}
\label{s:prior}
We obtain the segmentation prior $p( \labels, \z, \y | \nodes )$ by defining
\begin{equation}
  p( \labels, \z, \y, \g, \h | \nodes ) \propto \exp[-E( \labels, \z, \y, \g, \h | \nodes ) ]
  \nonumber
\end{equation}
with an energy
\begin{eqnarray}
  E( \labels, \z, \y, \g, \h | \nodes ) & = & E^z(\z,\h) + E^y(\y,\g) \\
  & & - \log q(\labels | \nodes) + \sum_i f( l_i, z_i, y_i ), \nonumber
  \label{e:energyjoint}
\end{eqnarray}
where $E^z(\z,\h)$ and $E^y(\y,\g)$ are the energy terms of two cRBMs that model tumor shape in $\z$ and $\y$, respectively, and $q(\labels | \nodes)$ is a deformable atlas that models the spatial configuration of the normal labels in $\labels$. Additionally, we use a restriction function defined as
\begin{equation}
  f(l,z,y) = 
  \begin{cases}
    \infty & \text{if } z=0 \text{ and } y=1 \\
    \infty & \text{if } z=1 \text{ and } l \notin B \\
    0 & \text{otherwise}
  \end{cases}.
  \label{e:rest}
\end{equation}
This function encodes that a core voxel can never appear outside the tumor-affected region $\mathbf{z}$, and that a tumor-affected voxel can never appear outside the brain. Note that it is only this restriction function that ties the labels $\labels$, $\z$, and $\y$ to each other. Without it, the segmentation prior would simply devolve into $p( \labels, \z, \y | \nodes ) = p( \labels | \nodes ) p( \z ) p( \y )$.

We will now present the two types of models that are included in this prior: the cRBMs on tumor shape in Section \ref{s:rbm}, and the atlas on the spatial configuration of normal head structures in Section \ref{s:atlas}. 

\subsubsection{Prior on tumor shape using cRBMs}
\label{s:rbm}
In order to model the spatial configuration of tumor tissue, we use cRBMs --
neural networks that can be interpreted as MRFs encoding
high-order interactions among voxels (``visible units'') through local connections to 
latent variables
(``hidden units'')
\citep{fischer2014}. In contrast to a standard
restricted Boltzmann machine \citep{smolensky1986information, freund1992unsupervised, hinton2002training}, where arbitrary weights can be assigned between the visible and the hidden units,
the weights of the connections in a cRBM
are in the form of filters that are much smaller than the image size and that are shared among all locations in the image \citep{lee2009}. This allows us to infer over large images without a predefined size.
We now present the model in only 1D for the sole purpose of avoiding cluttered equations, but it directly generalizes to 3D images.

\begin{figure}[H]
\begin{center}
\includegraphics[width=0.55\columnwidth]{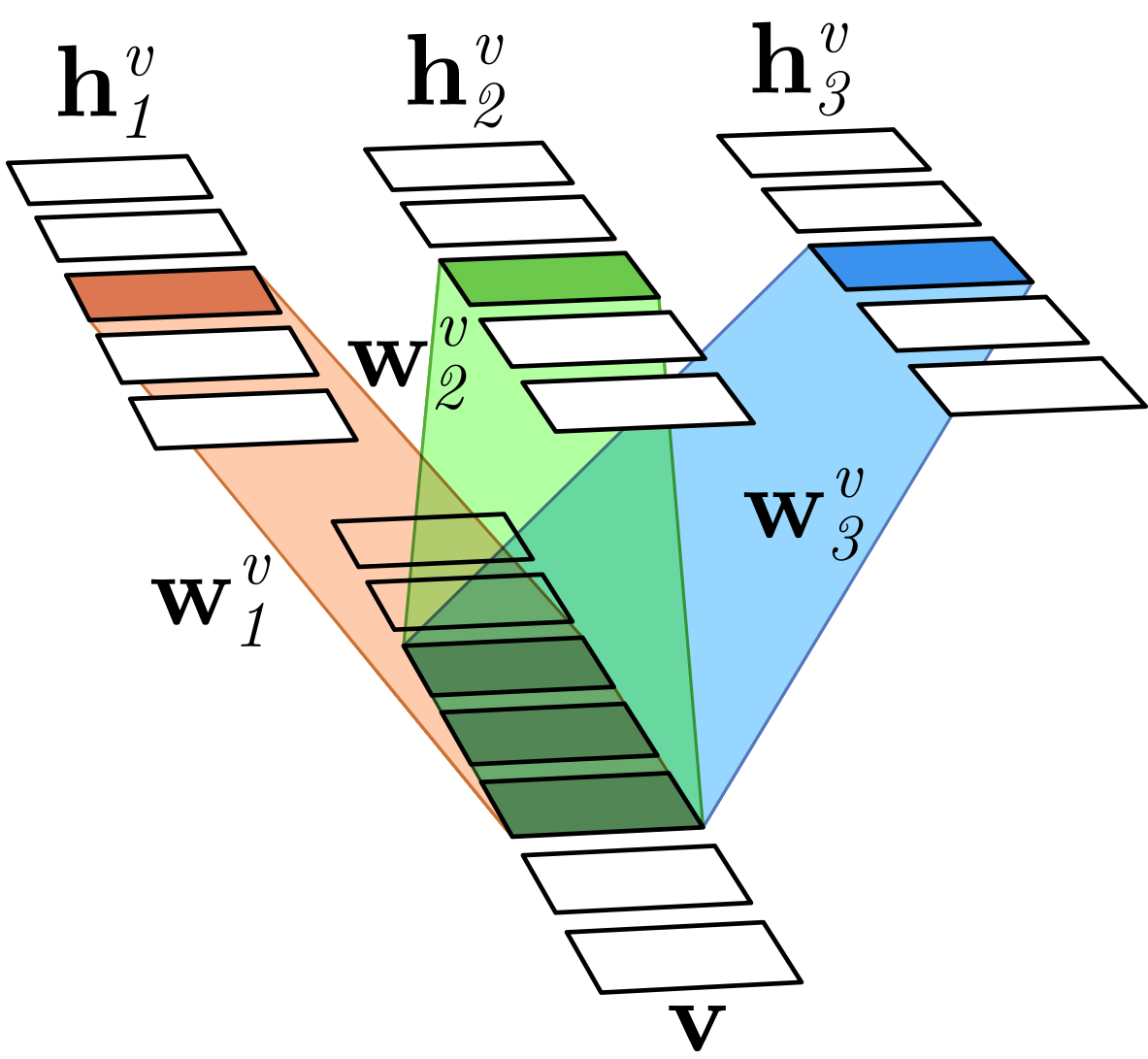}
\caption{A small 1D example of a cRBM with $\vis = (v_1, ..., v_7)$ and $\hid= \{\mathbf{h}^v_m\}_{m=1}^3$. Visible units (image voxels) are connected to hidden units in a hidden group $\mathbf{h}^v_m$ through a convolutional filter $\mathbf{w}^v_m$ of size 3. All locations in $\vis$ share the same filter weights. The connections are exemplified by the three central visible units which are connected to the central hidden unit in each group.}
\label{f:rbm}
\end{center}
\end{figure}

The distribution over visible units $\vis$ in a cRBM is defined as \begin{equation}
\label{e:pv}
p(\vis) = \sum_{\hid} \exp [-E^v(\vis,\hid)]
\end{equation}
with the energy term \citep{lee2009}
\begin{equation}
  E^v(\vis,\hid) = - \sum_{m=1}^M \mathbf{h}^v_m \bullet (\mathbf{w}^v_m \ast \vis) - \sum_{m=1}^M b^v_{m} \sum_{j=1}^J h^v_{mj} - a^v \sum_{i=1}^I v_i,\nonumber
\end{equation}
where $\hid = \{\mathbf{h}^v_m\}_{m=1}^M $ contains $M$ hidden groups, $\bullet$ denotes element-wise product followed by summation, and $\ast$ denotes spatial convolution. Each hidden group $\mathbf{h}^v_m$ is connected to the visible units in $\mathbf{v}$ with a convolutional filter $\mathbf{w}^v_m$ of size $r$, and contains $J=I-r+1$ hidden units. The filter $\mathbf{w}^v_m$ models interactions between the hidden and visible units, effectively detecting specific features in $\vis$. Furthermore, each hidden group has a bias $b^v_{m}$ and visible units have a bias $a^v$. These bias terms encourage units to be enabled or disabled when set to non-zero values. A small example of a cRBM can be seen in Figure \ref{f:rbm}. 

The computational appeal of this model is that no direct connections exist between two visible units or two hidden units, so that the visible units are independent of each other given the state of the hidden ones, and vice versa:
\vspace{0mm}\\
\begin{equation}
  p( \vis | \hid ) = \prod_i p( v_i | \hid ) 
  \quad \mathrm{and} \quad
  p( \hid | \vis ) = \prod_m \prod_j p( h^v_{mj} | \vis )
  \label{e:hidz}
\end{equation}\\
\vspace{-15mm}\\
\begin{equation}
  \mathrm{with}
  \quad
  p(v_{i} | \hid) \propto \exp \left[ v_i \left( \sum_m (\tilde{\mathbf{w}}^v_m \ast \mathbf{h}^v_m)_i + a^v \right) \right]
  \nonumber
\end{equation}
\begin{equation}
  \mathrm{and}
  \quad
  p(h_{mj}^v | \mathbf{v}) \propto \exp \left[ h_{mj}^v \left( \left(\mathbf{w}^v_m \ast \mathbf{v} \right)_j + b^v_{m} \right) \right],
  \nonumber
\end{equation}
where $\tilde{\mathbf{w}}$ denotes a mirror-reversed version of the filter $\mathbf{w}$. Although no direct connections exist among visible units, high-order connections are still obtained among them through the connections to the hidden units. This can be seen clearly by summing out the hidden units in Eq. \ref{e:pv} analytically \citep{fischer2014}, which gives us 
$p(\vis) \propto \exp [-E^v(\vis)]$ with
\begin{equation}
E^v (\vis) = \sum_{i=1}^{I-r+1} g (\vis_{i:i+r-1}) - a^v \sum_{i=1}^{I} v_i,
\end{equation}
where $i:i'$ denotes elements from $i$ to $i'$, \linebreak ${\text{and} \; g (\vis_{i:i+r-1}) = -\sum_m \log\left[1 + \exp(\mathbf{w}_m^v \bullet \vis_{i:i+r-1} + b_m^v)\right]}$ is a high-order MRF clique potential defined over groups of visible units as large as the filter size $r$. 
This can be contrasted to traditionally used MRF models for brain lesion shape, e.g., \citep{van1999automated,menze2010}, where $a^v$ is set to zero and the clique potentials are only between pairs of voxels in $\vis$, i.e., $r=2$, defined as $g(\vis_{i:i+1}) = \beta^v | v_i - v_{i+1} |$, where $\beta^v$ is a user-tunable hyperparameter.

In this paper, we use two separate binary cRBMs: one that models shape in the tumor-affected map $\z$ and one that models shape in the core map $\y$, with energies $E^z(\z,\h)$ and $E^y(\y,\g)$, defined exactly as for $\vis$. 
We learn suitable values for the filters and biases of these cRBMs by stochastic gradient descent on the log-likelihood using expert segmentations obtained from training data, 
as detailed
in Section \ref{sec:tumortrain}.

\subsubsection{Atlas-based prior on normal head structures}
\label{s:atlas}
To model the spatial configuration of normal head structures $q(\mathbf{l}|\boldsymbol \eta)$, we use the type of 
probabilistic atlas 
introduced in~\citep{VanLeemputTMI2009} and further validated in \citep{puonti2016}. 
It
is based on a 
deformable
tetrahedral mesh, where the parameters $\nodes$ are the spatial positions of the mesh nodes and $p(\boldsymbol \eta)$ is a topology-preserving deformation 
prior \citep{ashburner2000}. 
Each mesh node in the atlas 
is associated with a probability vector containing the probabilities of the $K$ normal head structures to occur at that node;
for a given mesh deformation,  
these vectors are interpolated using baricentric interpolation to yield probabilities  
$\pi_i(k|\boldsymbol \eta)$ for each structure $k$ in all voxels $i$.
Assuming
that structure labels at different voxels are conditionally independent given the node positions, this finally yields
\vspace{-2mm}
\begin{equation}
\nonumber
q(\mathbf{l}|\boldsymbol \eta) = \prod_{i=1}^I \pi_i(l_i|\boldsymbol \eta).
\end{equation}  

\begin{figure}[H]
\begin{center}
\begin{minipage}{0.59\columnwidth}
\includegraphics[width=1\columnwidth]{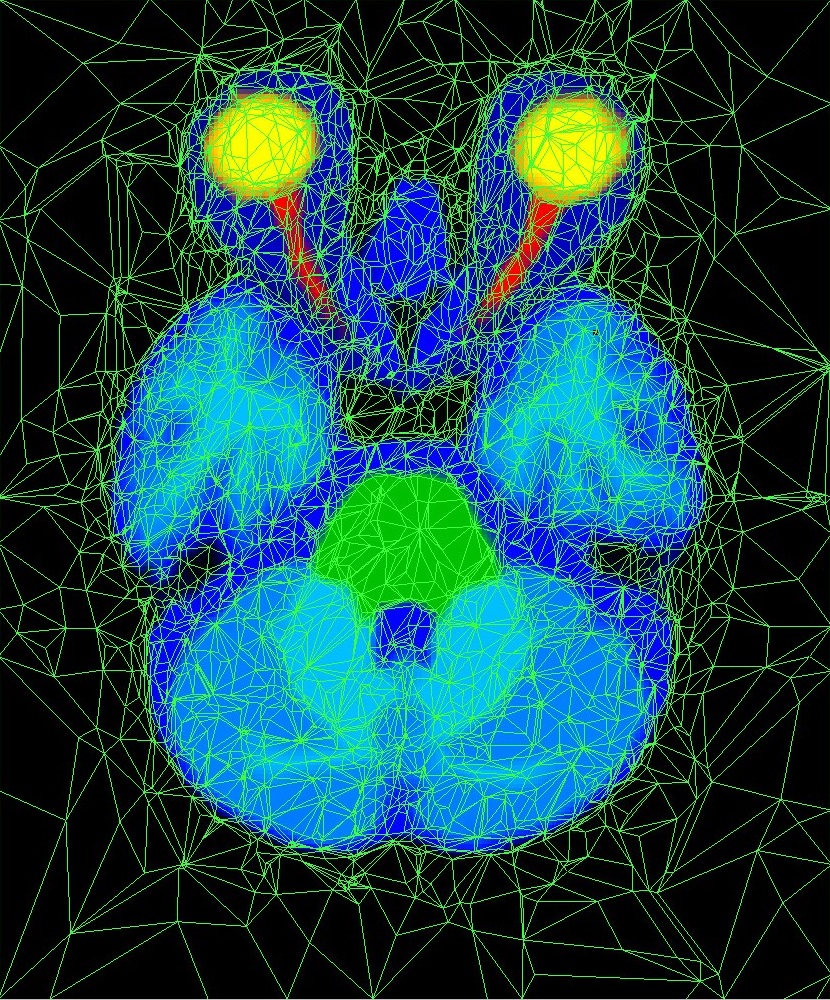}
\end{minipage}
\begin{minipage}{0.382\columnwidth}
\includegraphics[width=1\columnwidth]{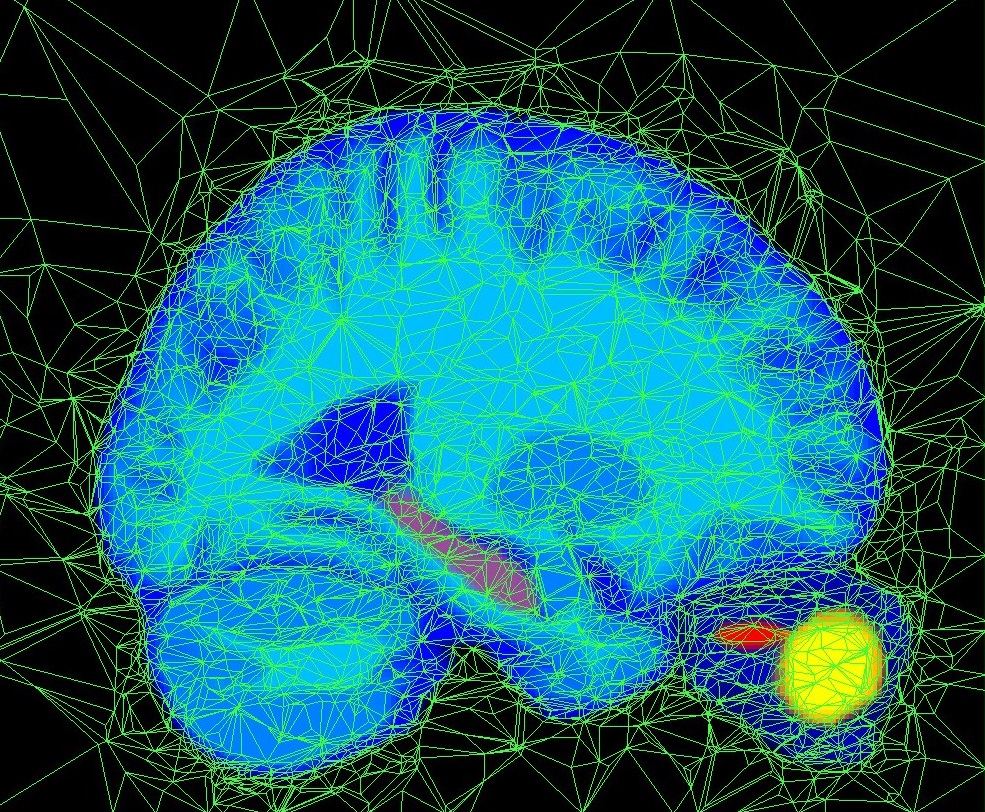} \\
\vspace{-3mm}
\includegraphics[width=1\columnwidth]{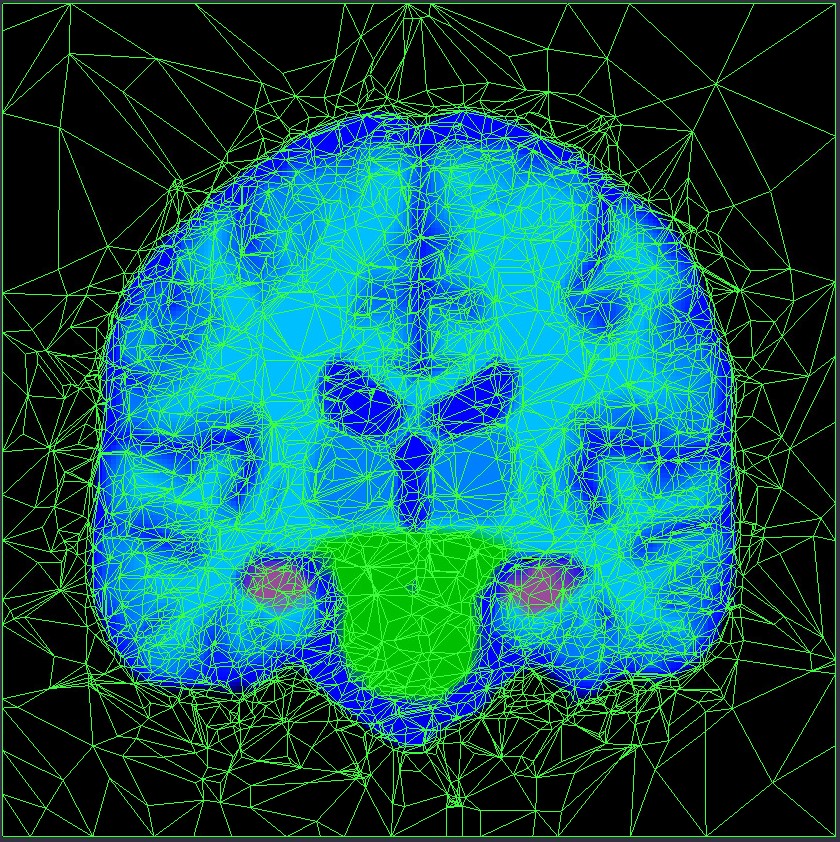}
\end{minipage}
\caption{The built atlas in axial, sagittal, and coronal view; shown in atlas space. Nodes and connections between nodes are shown in light green and probabilities of normal labels, interpolated between the nodes, are shown in varying colors (yellow = eye fluid, orange = eye tissue, red = optic nerves, green = brainstem, lilac = hippocampi, shades of blue = other normal labels).}
\label{f:atlas}
\end{center}
\end{figure}

As described in~\citep{VanLeemputTMI2009}, the atlas can be trained by a non-linear, group-wise registration of expert segmentations obtained from training data. The node positions in atlas space with associated label probabilities are optimized during this training process, as well as the topology of the mesh, where the mesh resolution adapts to be sparse in large uniform regions and dense at label borders. Figure \ref{f:atlas} shows the atlas that we built for the current paper; more details will be given in Section \ref{sec:atlastrain}. 

\subsection{Likelihood}
\label{s:likeh}
To link the labels $\labels$, $\z$ and $\y$ to image intensities, we use $X=12$ Gaussian mixture models (GMMs) in the likelihood function $p(\mathbf{D}|\mathbf{l},\mathbf{z},\mathbf{y}, \param)$, where each GMM models the intensity distribution of certain label combinations. Some GMMs are connected to several label combinations, e.g., left and right hippocampus are modeled by the same GMM as both hippocampi have the same intensity properties, and any voxel $i$ that belongs to edema (i.e., $z_i=1$, $y_i=0$ and $l_i \in B$) is modeled by a single GMM. 
In order to map a voxel $i$ with $l_i$, $z_i$ and $y_i$ to a specific GMM, we therefore introduce a mapping function $x(l_i,z_i,y_i)$, which is detailed in Table \ref{t:map}.
Additionally, we model so-called bias fields that typically corrupt MR scans as additive effects by linear combinations of spatially smooth basis functions. A bias field is a multiplicative low-frequency imaging artifact, so to model it as an additive effect we work with log-transformed intensities throughout this paper, as in \citep{wells1996adaptive,leemput1999b}.

\begin{table}
\caption{Mapping function $x(l,z,y)$ that maps combinations of $l$, $z$ and $y$ to GMMs. Note that combinations $\{z=1, \forall y, l \notin B\}$ and $\{z=0, y=1, \forall l\}$ will never occur due to the restriction function in Eq. \ref{e:rest}.
}
\label{t:map}
\footnotesize
\center
\begin{tabular}{l l}
\hline
\vspace*{-3mm} \\
Combinations of $l, z,$ and $y$ & $x(l,z,y)$ \\
\hline
$z = 1, y=1, \text{ and } l \in B$ & core \\
$z = 1, y=0, \text{ and } l \in B$ & edema \\
$z = 0, y=0, \text{ and } l \in$ \\
\quad \{GM, L/R hippocampus\} & global gray matter (GGM)\\
\quad \{WM, brainstem\} & global white matter (GWM)\\
\quad \{L/R optic nerve, L/R eye tissue\} & global nerves/eye tissue (GNE)\\
\quad \{L/R eye fluid\} & global eye fluid\\
\quad   CSF & CSF\\
\quad   background & background\\
\quad  unspecified brain tissue & unspecified brain tissue \\
\quad   optic chiasm & optic chiasm\\
\quad   eye socket fat  & eye socket fat\\
\quad   eye socket muscles & eye socket muscles\\
\hline
\end{tabular}
\end{table}

Specifically, we define the likelihood function as
\begin{equation}
p(\mathbf{D}|\mathbf{l},\mathbf{z},\mathbf{y}, \param) = \prod_i p_i(\mathbf{d}_i | x(l_i,z_i,y_i), \param)
\nonumber
\end{equation}\\
\vspace{-8mm}
\begin{equation}
\nonumber
\text{with} \quad
p_i(\mathbf{d}_i | x, \param) = \sum_{g=1}^{G_x} \gamma_{xg} \mathcal{N}(\mathbf{d}_i |\boldsymbol \mu_{xg} + \mathbf{C} \boldsymbol \phi_i, \boldsymbol \Sigma_{xg}),
\end{equation}
where $\mathcal{N} (\mathbf{d} |\boldsymbol \mu, \boldsymbol \Sigma)$
denotes a multivariate normal distribution with mean $\boldsymbol \mu$ and covariance $\boldsymbol \Sigma$; 
$G_x$ is the number of components in the $x$th GMM; and $\gamma_{xg}$, $\boldsymbol \mu_{xg}$ and $\boldsymbol \Sigma_{xg}$ are the weight, mean and covariance matrix of component $g$. The weights satisfy the constraints $\gamma_{xg} \geq 0$ and $\sum_{g=1}^{G_x} \gamma_{xg} = 1$. Furthermore, the bias fields corrupting MR scans are modeled by $\boldsymbol \phi_i$ and $\mathbf{C}$. The column vector $\boldsymbol \phi_i \in \mathbb{R}^P$ evaluates $P$ spatially smooth basis functions at voxel $i$ and $\mathbf{C} = (\mathbf{c}_1, ...,\mathbf{c}_N)^T$ denotes the parameters of the bias field model, where $\mathbf{c}_n \in \mathbb{R}^P$ are the parameters for image contrast $n$. Finally, all likelihood parameters are jointly collected in $\param = \{ \{ \gamma_{xg}, \boldsymbol \mu_{xg}, \boldsymbol \Sigma_{xg}\}\forall xg , \mathbf{C} \}$.

We use a restricted conjugate prior $p (\boldsymbol \theta)$ on the likelihood parameters: 
\begin{equation}
\label{e:thetaprior}
  p( \boldsymbol \theta ) 
  \propto 
  \begin{cases}
  \prod_x 
  \left[
    \mathrm{Dir} (\boldsymbol \gamma_x | \alpha_0) 
    \prod_{g=1}^{G_x} \mathrm{IW} (\boldsymbol \Sigma_{xg} | \upsilon_x^{0}, \mathbf{S}_x^{0}) 
  \right]  
  \\
  \quad \;\; \text{if constraints on $\{\boldsymbol \mu_{xg}\}$ are satisfied} \\
  0 \quad \text{otherwise},
  \end{cases}
\end{equation}
where
we 
have used uniform priors on the bias field parameters $\mathbf{C}$ and the mean vectors $\{ \boldsymbol \mu_{xg} \}$, and conjugate priors on the covariance matrix of each component and mixture weights of each GMM following the definitions in \citep{murphy2012}.
To regularize the covariance matrices, we use inverse-Wishart distributions $\mathrm{IW} (\boldsymbol \Sigma | \upsilon_x^{0}, \mathbf{S}_x^{0})$, where $\mathbf{S}_x^{0}$ is a prior scatter matrix with strength $\upsilon_x^{0}$; and, to discourage removal of components, we use symmetric Dirichlet distributions $\mathrm{Dir} (\boldsymbol \gamma | \alpha_0)$, where prior strength  $\alpha_0$ is non-zero.
Additionally, we add certain linear constraints on $\{\boldsymbol \mu_{xg}\}$ to encode prior knowledge about overall tumor appearance relative to normal brain tissue in typical MR sequences for brain tumor imaging. 
Tuning of the likelihood function and its parameter prior is detailed in Section \ref{sec:liktrain}. 

\subsection{Inference}
\label{s:infer}
Exact inference of the posterior $p(\mathbf{l},\mathbf{y},\mathbf{z}|\mathbf{D})$ in Eq. \ref{e:segmpost} is computationally
intractable because it marginalizes over all of the uncertainty in the model parameters and the hidden units of the cRBM models. We therefore resort to Markov chain Monte Carlo (MCMC) 
techniques to sample from all unknown variables (except the atlas node positions $\boldsymbol \eta$, as detailed below), followed by voxel-wise majority voting on the segmentation samples to obtain the final segmentation. 
This procedure is detailed in Section \ref{s:sampler}.

Although it is possible to also sample from  $\boldsymbol \eta$, as shown in~\citep{iglesias2013},
this is considerably more computationally expensive
and was not implemented
in this paper.
Instead, we ignore the uncertainty on deformations and use a suitable point estimate of the
atlas node positions $\boldsymbol{ \hat{\eta}}$ obtained with a simplified model, which we describe in Section \ref{s:init}. We also obtain an initial state of the sampler from this simplified model. 

\subsubsection{MCMC sampler}
\label{s:sampler}
Given a point estimate of the atlas node positions $\hat{\boldsymbol \eta}$, we generate samples of the labels $\labels$, $\z$ and $\y$ from  $p(\mathbf{l},\mathbf{z},\mathbf{y}|\mathbf{D},\hat{\boldsymbol \eta})$ by sampling from 
$p(\mathbf{l},\mathbf{z},\mathbf{y}, \h, \g, \boldsymbol \theta |\mathbf{D},\hat{\boldsymbol \eta})$
using a blocked Gibbs sampler, and discarding the samples of $\g, \h$ and $\boldsymbol \theta$. The sampler, which is illustrated in Algorithm \ref{a1}, 
iteratively draws
each set of variables from its conditional distribution given the other variables;
with the exception of $\boldsymbol \theta$ 
this is straightforward to implement
as each conditional distribution factorizes over its components. The hidden units $\h$ and $\g$ are sampled as in Eq. \ref{e:hidz}, and the labels are sampled from
\begin{equation}
  p( \mathbf{l},\mathbf{z},\mathbf{y} | \mathbf{D}, \h, \g, \boldsymbol \theta, \hat{\boldsymbol \eta})
  = \prod_i p_i( l_i,z_i,y_i | \mathbf{d}_i, \h, \g, \boldsymbol \theta, \hat{\boldsymbol \eta})
  \label{e:segmcond}
\end{equation}\vspace{-5mm}\\
with\\
\vspace{-5mm}
\begin{multline*}
  p_i( l_i,z_i,y_i | \mathbf{d}_i, \h, \g, \boldsymbol \theta, \hat{\boldsymbol \eta})
  \propto
  \\
  \quad
  \begin{array}{l}
  p_i( \mathbf{d}_i | l_i,z_i,y_i, \boldsymbol \theta )
  \pi_i( l_i )
  \exp\left[ 
    z_i \left( \sum_m (\tilde{\mathbf{w}}^z_m \ast \mathbf{h}^z_m)_i + a^z \right)
  \right]  
  \\
  \qquad
  \exp\left[ 
    y_i \left( \sum_m (\tilde{\mathbf{w}}^y_m \ast \mathbf{h}^y_m)_i + a^y \right) 
  \right]
  \exp\left[ 
    -f(l_i,z_i,y_i)
  \right]
  .
  \end{array}
\end{multline*}
Sampling from the 
conditional distribution $p( \boldsymbol \theta | \data, \mathbf{l},\mathbf{z},\mathbf{y})$
is more difficult due to interdependencies among the various components of $\boldsymbol \theta$, and is detailed in \ref{a:step2}. 

We obtain the final estimate of the labels  $\hat{\labels}, \hat{\z}$, and $\hat{\y}$ by voxel-wise majority voting, separately on each variable, over $S$ collected samples after an initial burn-in period of $S_\text{burn-in}$ samples. 

\begin{algorithm}
\caption{ MCMC sampler to obtain  $\hat{\labels}, \hat{\z}, \hat{\y}$}
\label{a1}
\begin{tabular}{l}
\textbf{Input:} $\labels^{(0)}, \z^{(0)}, \y^{(0)}, \hat{\nodes}$\\
\textbf{Output} final estimates of labels $\hat{\labels}, \hat{\z}, \hat{\y}$\\
\textbf{for} $s = 1$ to $(S_\text{burn-in} + S)$\\
\quad Sample $\param$ from $p(\param|\data, \labels^{(s-1)},\z^{(s-1)},\y^{(s-1)})$, \\
\quad\quad\ detailed in \ref{a:step2}\\
\quad Sample $\h$ from $p(\h|\z^{(s-1)})$, see Eq. \ref{e:hidz}\\
\quad Sample $\g$ from $p(\g|\y^{(s-1)})$, see Eq. \ref{e:hidz}\\
\quad Sample $\labels^{(s)}, \z^{(s)}, \y^{(s)}$ from \\
\quad\quad\ $p( \mathbf{l},\mathbf{z},\mathbf{y} | \mathbf{D},\h, \g, \boldsymbol \theta, \hat{\boldsymbol \eta})$ in Eq. \ref{e:segmcond}\\
\textbf{end for} \\
Final $\hat{\labels}, \hat{\z}, \hat{\y}$ obtained by voxel-wise majority voting \\ of samples in $\{\labels^{(s)},\z^{(s)},\y^{(s)}\}_{s=S_\text{burn-in} + 1}^{S_\text{burn-in} + S}$\\
\end{tabular}
\end{algorithm}

\subsubsection{Simplified model to obtain atlas node position estimates and initial state of sampler}
\label{s:init}
For the purpose of estimating appropriate
atlas node positions $\boldsymbol{\hat{\eta}}$ and to obtain an initial state $\{\labels^{(0)}, \z^{(0)}, \y^{(0)}\}$ for the MCMC sampler, we use a simplified model in which the 
non-local dependencies among the voxels introduced by the cRMB shape models are removed. 
In particular, we set the filter weights $\{\mathbf{w}^z_m\}_{m=1}^M$ and $\{\mathbf{w}^y_m\}_{m=1}^M$ to zero values, effectively removing the hidden units from the model, and set the visual bias values so that a fraction $w=0.1$ of normal voxels is expected to be tumorous, and within these voxels a fraction $u=0.5$ is expected to be tumor core. We achieve this by setting the visual biases $a_y = \log(\frac{u}{1-u})$ and $a_z = \log( \frac{w-wu}{1 - w})$. 
This reduces the model to the same form as in~\citep{puonti2016}, and we can therefore use the same approach for optimization,
i.e., 
by alternating between optimizing the likelihood parameters $\boldsymbol \theta$ with a generalized expectation-maximization (GEM) algorithm \citep{dempster1977} and optimizing the atlas node positions $\boldsymbol \eta$ with a general-purpose gradient-based optimizer.
Algorithm \ref{a2} illustrates this approach, which is implemented as in \citep{puonti2016} 
with a few exceptions. In particular, for the atlas node positions a more efficient optimizer is used (limited-memory BFGS \citep{Liu1989}). Furthermore,
the linear constraints in the prior $p(\param)$ (Eq.~\ref{e:thetaprior}) alter the relevant update equations in the GEM algorithm, which is detailed in \ref{a:step1}.
Finally, 
as in \citep{puonti2016}, all Gaussian component parameters in $\param$ are initialized based on the atlas prior after affine registration, except the mean values for the tumor GMMs. These are instead initialized based on prior knowledge about overall tumor appearance in typical MR sequences for brain tumor imaging, as detailed in Section \ref{sec:liktrain}. 

After convergence of the parameter optimization with this simplified model, we record $\hat{\boldsymbol \eta}$ 
and 
compute the 
\textit{maximum a posteriori} segmentation 
\begin{eqnarray}
  \{\labels^{(0)},\z^{(0)},\y^{(0)}\} 
  & = & \arg\max_{\labels,\z,\y} p(\mathbf{l},\mathbf{y},\mathbf{z}|\mathbf{D}, \hat{\boldsymbol \theta}, \hat{\boldsymbol \eta}) 
  \nonumber\\
  & = & \arg\max_{\{l_i,z_i,y_i\}} \prod_i p( l_i, z_i, y_i | \mathbf{d}_i, \hat{\boldsymbol \theta}, \hat{\boldsymbol \eta} ),
  \nonumber
\end{eqnarray}  
which is used as the initial state for the MCMC sampler.

\begin{algorithm}
\caption{ Initial algorithm to obtain $\labels^{(0)}, \z^{(0)}, \y^{(0)}, \hat{\nodes}$}
\label{a2}
\begin{tabular}{l}
\textbf{Input:} $\data$, initial affine transformation of atlas $\hat{\nodes}$\\
\textbf{Output: $\labels^{(0)}, \z^{(0)}, \y^{(0)}, \hat{\nodes}$}\\
Change tumor prior to a simplified version\\
Initialize $\hat{\param}$ \\
\textbf{until} convergence\\
\quad Optimize $\hat{\param} = \arg \max_{\param} p(\param | \data, \hat{\nodes})$\\
\quad Optimize $\hat{\nodes} = \arg \max_{\nodes} p(\nodes | \data, \hat{\param})$\\
\textbf{end until}\\
Record $\hat{\nodes}$\\
Compute \textit{maximum a posteriori} segmentation\\
$\{\labels^{(0)}, \z^{(0)}, \y^{(0)}\} = \arg \max_{\labels,\z,\y} p(\labels,\z,\y|\data,\hat{\param},\hat{\nodes})$\\
\end{tabular}
\end{algorithm}

\section{Training and tuning of the model}
\label{sec:train}
In this section, we describe how we trained the deformable atlas $q(\labels | \nodes)$ (in Section \ref{sec:atlastrain}) and the two cRBMs modeling $\z$ and $\y$ (in Section \ref{sec:tumortrain}), which together make up the segmentation prior in our model. Furthermore, we describe overall tuning of the method
in Section \ref{sec:liktrain}.   

To train the deformable atlas, we used the same training dataset as in \citep{puonti2016}, which is also the training data of the publicly available software package FreeSurfer \citep{fischl2012}. This dataset consists of 39 subjects (without any tumors) with dozens of neuroanatomical structures within the brain segmented by experts, following
a validated semi-automated protocol developed at the Center for Morphometric Analysis (CMA), MGH, Boston \citep{caviness1989, caviness1996, kennedy1989}. We call this dataset \textit{the atlas training dataset}.

For all other parts of the model, we used the training dataset of the brain tumor segmentation (BRATS) challenge that was held in conjunction with the BrainLes workshop at the 2015 MICCAI conference. This dataset consists of 220 high-grade gliomas and 54 low-grade gliomas of varying types, with publicly available ground truth segmentations of tumor, which include annotations of four tumor regions: edema and three regions inside tumor core. 30 subjects were manually segmented (20 high-grade, 10 low-grade), while the rest have fused segmentations from highly ranked algorithms from previous editions of the BRATS challenge. The included MR sequences are T2-weighted FLAIR (2D acquisition), T2-weighted (2D acquisition), T1-weighted (2D acquisition), and T1-weighted with contrast enhancement (T1c, 3D acquisition). The scans have been acquired at different centers, with varying magnetic field strength and resolution. All data were resampled to 1 mm iso\-tropic resolution by the challenge organizers. We call this dataset \textit{the BRATS 2015 training dataset}.

\subsection{Training the deformable atlas}
\label{sec:atlastrain}
We automatically trained the tetrahedral mesh atlas, shown in Figure \ref{f:atlas} and described in Section \ref{s:atlas}, from expert segmentations from the atlas training dataset. We emphasize that only the manual segmentations are needed for this purpose, and that the intensity information of the original MR scans from which these were derived was not used.

As we are specifically interested in structures applicable to radiation therapy, we merged
some of the manually segmented structures into larger labels before building the atlas. 
Specifically, we kept the segmentations for the OARs \textit{brainstem}, \textit{optic chiasm} and left and right \textit{hippocampus}; as well as the background label. We merged all other structures into the following catch-all labels: cerebrospinal fluid (CSF), and remaining white matter (WM) and gray matter (GM). Two important OARs were not included in the available expert segmentations, as they are located outside of the brain -- namely \textit{optic nerves} and  \textit{eyes}. We therefore performed additional manual delineations for the left and right structures of these two extra OARs. To provide some context around these structures, we also delineated the muscles and fat in the eye sockets into two separate labels. We further separated the left and right eye into two labels each: \textit{eye fluid} describing the fluid and gel inside an eye and \textit{eye tissue} describing the lens and the solid outer layer of an eye.  

To build the atlas, we chose the resulting segmentations of a representative subset of 10 subjects. We selected 10 subjects as manual delineations are time consuming and we have previously shown that adding more subjects does not substantially increase 
the average segmentation performance \citep{puonti2016}. After building the atlas, we added an unspecified brain tissue label designed to capture normal structures that are not specified in the atlas, such as blood vessels. Towards this end, we added a constant prior probability of 0.01 for this label in each mesh node's probability vector and re-normalized the probability vector to ensure that the values sum to one. Overall, we use $K=17$ normal head structure labels, listed in Table \ref{t:lab}.

\subsection{Training the cRBMs}
\label{sec:tumortrain}
To learn suitable values for the filters and biases of the cRBMs modeling $\z$ and $\y$, described in Section \ref{s:rbm}, we used the 30 \textit{manual} tumor segmentations from the BRATS 2015 training dataset, again without using any associated intensity information. As the number of segmentations is small, we augmented the dataset by flipping the segmentations in eight different directions, yielding a dataset of 240 tumor segmentations. To form binary segmentations corresponding to $\z$ and $\y$, we merged tumor regions in the manual segmentations: all four regions for $\z$ and the three tumor core regions for $\y$. We learned the filters and bias terms through stochastic gradient ascent on the 
log-probability of the tumor segmentations under the cRBM model.
To efficiently approximate the 
gradients, we used the contrastive divergence (CD) approximation with one block-Gibbs sampling step \citep{hinton2002training}
together with the so-called enhanced gradient which has been shown to improve learning \citep{cho2011,melchior2013}. Each cRBM was trained with 9600 gradient steps of size 0.1. A subset of 10 randomly selected segmentations (a so-called mini-batch) was used to approximate the gradient at each step. 

\begin{figure}[H]
\begin{center}
\includegraphics[width=0.9\columnwidth]{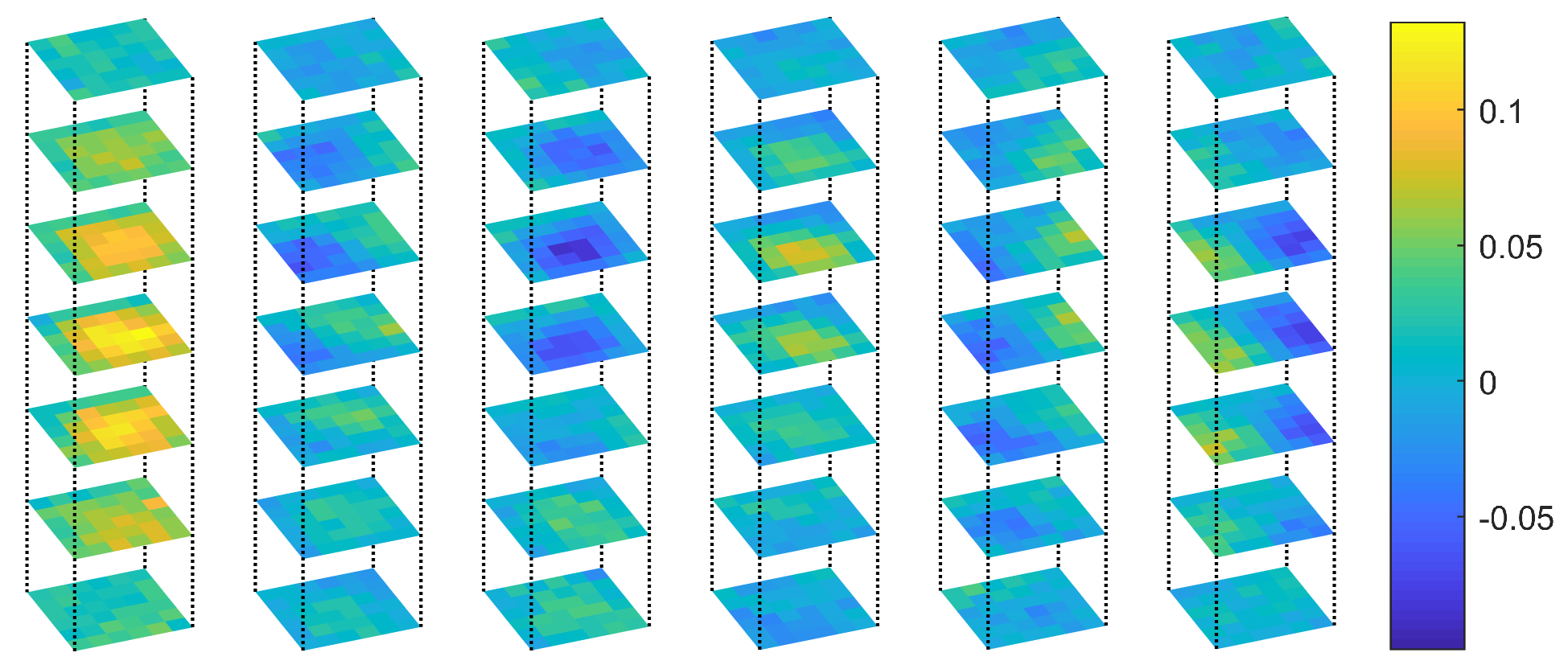}
\caption{A subset of six filters out of 40 learned for the tumor core cRBM. Slices of each filter are displayed from top to bottom.}
\label{f:filters}
\end{center}
\end{figure}

We used the same settings for both cRBMs. The filter size and number of filters were set by pilot experiments on a separate subset of the BRATS 2015 training dataset. Choosing a larger filter size would increase the number of parameters which may result in overfitting, while a smaller filter size might not capture long-range features. Empirically, we found that by tying neighboring parameters in a filter we can reduce the number of parameters while still capturing long-range features. Specifically, we tied filter parameters in $(2 \times 2 \times 2)$ blocks of voxels, effectively treating each block as one parameter. We used
$M=40$ filters of size $(14 \times 14 \times 14)$ (i.e., $7 \times 7 \times 7$ blocks) corresponding to 40 hidden groups. In our pilot experiments, this configuration performed better than other combinations of 20, 30 and 40 filters of sizes between 10 and 18. Figure \ref{f:filters} shows a subset of six filters learned for the core cRBM. 

\subsection{Tuning}
\label{sec:liktrain}
The tuning of the model described in this section is based on initial experiments on the full BRATS 2015 training dataset. 
We use $S=50$ samples from the MCMC sampler, after an initial burn-in period of $S_\text{burn-in} = 200$ (cf.~Algorithm \ref{a1}). 
In the likelihood function $p(\data|\labels,\z,\y,\param)$, described in Section \ref{s:likeh}, we associate three Gaussian components (i.e., $G_x=3$) with the GMMs of \textit{core, eye socket muscle}, and \textit{background}; two components with the GMMs of \textit{eye socket fat, CSF,} and \textit{GNE} (global optic nerves/eye tissue); and one component with all other GMMs. Additionally, we use the 64 lowest frequencies of the 3D DCT as bias field basis functions, i.e., $P = 64$.

In the likelihood parameter prior $p(\boldsymbol \theta)$ defined in Eq. \ref{e:thetaprior}, the linear constraints on 
the Gaussian means
$\{\boldsymbol \mu_{xg}\}$ were set by building statistics of their values in the BRATS 2015 training data. Specifically, we estimated
the average Gaussian mean values using automatic segmentations produced by our method, but with the tumor labels fixed to the ground truth. Based on the resulting statistics, we set constraints for the Gaussian mean values relating to edema and enhanced core in the MR sequences FLAIR and T1c. Enhanced core, which is the core region that is enhanced in T1c, is specifically targeted by setting constraints on only one of the Gaussian components associated with core. Additionally, we set constraints on the mean values relating to the unspecified brain tissue and optic chiasm as to ascertain that these labels will not interfere with the tumor segmentation. All constraints are in relation to the mean values of global WM (GWM) and global GM (GGM), and are shown in Table \ref{t:const}. Note that the image intensities are log-transformed, so an added logarithm of a value is equivalent to that value being multiplied by the original intensities.

For the inverse Wishart distribution in Eq. \ref{e:thetaprior}, we set 
the scatter matrix 
$\mathbf{S}_x^0 = \upsilon_x^0 X^{-2} \mathrm{diag}\left[ \sum_i(\mathbf{d}_i - \mathbf{\bar{d}})(\mathbf{d}_i - \mathbf{\bar{d}} )^T / I \right], \text{ with } \mathbf{\bar{d}} = \sum_{i} \mathbf{d}_{i} / I$
and strength $\upsilon_x^0 = N + 10^{-1} I_{x} /G_x$,
where $I_{x}$ is the expected number of voxels for each GMM, obtained from the atlas for normal structures and from the BRATS 2015 training data for tumor. Because the unspecified brain tissue label should catch any unspecified brain tissue, we use a wider scatter matrix for the GMM of this label, with $X$ set to 1.  Finally, we set $\alpha_0 = 1 + 10^{-4} I$ in the Dirichlet prior of Eq. \ref{e:thetaprior} for each GMM. 
\begin{table}
\caption{Constraints on mean values of Gaussian components.}
\vspace{-2mm}
\center
\footnotesize
\bgroup
\def\arraystretch{1.5}
\begin{tabular}{c c c}
\hline
\multicolumn{3}{c}{Edema (TE)}  \\
\hline
$\mu_{\,\text{TE}}^{\text{FLAIR}}$& $\geq$ & $ \max \left( \mu_{\, \text{GWM}}^{\text{FLAIR}}, \mu_{\, \text{GGM}}^{\text{FLAIR}} \right) + \log 1.15$\\
\hline
\multicolumn{3}{c}{Core, Gaussian component relating to enhanced core (denoted TC1)} \\
\hline
$\mu_{\, \text{TC1}}^{\text{FLAIR}}$ & $\geq$ & $\max \left( \mu_{\, \text{GWM}}^{\text{FLAIR}}, \mu_{\, \text{GGM}}^{\text{FLAIR}} \right)$\\
$\mu_{\, \text{TC1}}^{\text{T1c}}$ & $\geq$ & $\max \left( \mu_{\, \text{GWM}}^{\text{T1c}}, \mu_{\, \text{GGM}}^{\text{T1c}} \right) + \log 1.10$\\
\hline
\multicolumn{3}{c}{Unspecified brain tissue (US)}\\
\hline
$\mu_{\, \text{US}}^{\text{FLAIR}}$ & $\leq$ & $\min \left( \mu_{\, \text{GWM}}^{\text{FLAIR}}, \mu_{\, \text{GGM}}^{\text{FLAIR}} \right) - \log 1.05$\\
$\mu_{\, \text{US}}^{\text{T1c}}$ & $\leq$ & $\min \left( \mu_{\, \text{GWM}}^{\text{T1c}}, \mu_{\, \text{GGM}}^{\text{T1c}} \right) - \log 1.05$\\
\hline
\multicolumn{3}{c}{Chiasm (CH)}\\
\hline
$\mu_{\, \text{CH}}^{\text{FLAIR}}$ & $\leq$ & $\min \left( \mu_{\, \text{GWM}}^{\text{FLAIR}}, \mu_{\, \text{GGM}}^{\text{FLAIR}} \right)$\\
\hline
\end{tabular}
\egroup
\label{t:const}
\end{table}

\paragraph*{Initialization of the simplified model of Algorithm \ref{a2}}
As described in Section \ref{s:init}, all Gaussian component parameters are initialized based on the atlas prior, except the mean values associated with tumor. If the flat tumor prior in the simplified model of Algorithm \ref{a2} would be used, these mean values would be initialized as the average intensities within the brain, which are far away from typical tumor intensities. Therefore, we instead initialize these mean values based on the distance (measured in standard deviation) from the average data intensity in that image. Based on initial pilot experiments on the BRATS 2015 training data, we set these distances as in Table \ref{t:distances}.

\paragraph*{Specific settings for tumor core} 
The GMM connected to tumor core needs special care due to the flat tumor prior used in the simplified model of Algorithm \ref{a2}. Tumor core regions can vary widely in their intensity distribution and can also have a similar intensity distribution to edema and normal tissue. This fact creates challenges when estimating the parameters of the core GMM during inference in Algorithm \ref{a2}, as the flat tumor prior has no notion of tumor shape. The easiest region to recognize only by intensity is the region that is enhanced in T1c. Thus, we temporarily restrict all three Gaussian components associated with core to have identical mixture parameters while using the simplified model, and specifically target the enhanced region. We then release the restriction before starting the sampler (Algorithm \ref{a1}). Additionally, to help the full cRBM-based model to capture other core regions in the vicinity of the enhanced region, we randomly change a fifth of the edema voxels ($z_i^{(0)} = 1$ and $y_i^{(0)} = 0$) in the initial state to core voxels ($z_i^{(0)} = 1$ and $y_i^{(0)} = 1$).

\section{Experiments and results}
\label{sec:exp}
To evaluate our method, we conduct experiments on three different datasets from different imaging centers with varying input data, including CT images and several MR sequences. The varying input data enables us to assess our method's ability to handle images from different modalities, MR sequences and scanner platforms. In Section~\ref{sec:clin}, we test our method on a dataset of 70 glioblastoma patients that have undergone radiation therapy treatment at Rigshospitalet in Copenhagen, Denmark. We call this dataset \textit{the Copenhagen dataset}. It includes all data needed for a radiation therapy session, which enables us to test our method's performance on both tumor and OAR segmentation, as well as to conduct a dosimetric evaluation. In this dataset, we will also vary the input data to the method from the available images to test the effect this has on the segmentation performance. Furthermore, we will compare our cRBM-based method to that of the same method but instead using first-order MRFs. In Section~\ref{sec:brats}, we compare our method's performance on segmenting tumors to that of top-performing methods in the 2015 BRATS challenge, using the challenge's test dataset of 53 patients from varying centers, which we call the \textit{the BRATS 2015 test dataset}. Lastly, in Section~\ref{sec:dir}, we further test our method's ability to adapt to varying input data by using a dataset of seven patients with a different set of acquired images, including an MR sequence not present in the other datasets, scanned at the National Hospital for Neurology and Neurosurgery, UCLH NHS Foundation Trust, London, UK. We call this dataset \textit{the London dataset}. 

\begin{table}
\caption{Distances in standard deviations, to initialize tumor GMMs.}
\label{t:distances}
\center
\footnotesize
\vspace{-2mm}
\begin{tabular}{l | c c c c}
$x$  & FLAIR & T2 & T1 & T1c\\
\hline
core & 1 & 0.7 & 0.2 & 1.5\\
edema  & 1 & 0.7 & 0.2 & 0.2 \\
\end{tabular}
\end{table}

Throughout this section, we employ two widely used metrics -- Dice score and Hausdorff distance -- to compare our\linebreak method's segmentations to the manual segmentations in the \linebreak datasets.
A Dice score measures overlap between two segmentations, where a score of zero means no overlap and a score of one means a perfect overlap. In contrast, a Hausdorff distance evaluates the distance between the surfaces of two segmentations. As in the BRATS challenges, we use a robust version of this metric. A further description of these two metrics can be found in the BRATS reference paper~\citep{menze2015}. 

The entire algorithm was implemented in MATLAB 2015b, except for the atlas mesh deformation which was implemented in C++. Segmenting one subject takes around 40 minutes on a Core i7-5930K CPU with 32 GB of memory, with roughly equal time spent on Algorithm \ref{a1} and \ref{a2} described in Section \ref{s:infer}. 

\subsection{Results for Copenhagen dataset}
\label{sec:clin}
To evaluate our method's performance on segmenting both OARs and tumors, we use the Copenhagen dataset, which consists of 70 glioblastoma patients that have undergone radiation therapy treatment at Rigshospitalet in Copenhagen, Denmark, in 2016 (GTV size range: 5-205 cm$^3$). As part of their radiation therapy workup, these patients have been scanned with a CT scanner and a Siemens Magnetom Espree 1.5T MRI scanner. The dataset includes three MR sequences: T2-weighted FLAIR \linebreak (transversal 2D-acquisition), T2-weighted (T2, transversal 2D-acquisition) and T1-weighted with contrast enhancement (T1c, 3D-acquisition); with a voxel size of ($1 \times 1 \times 3$), ($1 \times 1 \times 3$) and ($0.5 \times 0.5 \times 1$) mm$^3$ respectively. The CT scans have a voxel size of ($0.5 \times 0.5 \times 1$) mm$^3$. As part of the treatment planning, the GTV (corresponding to tumor core) and several OARs (including hippocampi, brainstem, eyes, optic nerves and chiasm) have been manually delineated in CT-space, with the MR sequences transformed to this space. As the only pre-processing step for our method, we co-register the MR and CT scans and resample them to 1 mm isotropic resolution.

\subsubsection{Evaluation of results on three data combinations} 
To test the ability of our method to adapt to varying input data, we evaluate the segmentation results obtained with three different data combinations. In the first combination, we use all available data, i.e., \{T1c, FLAIR, T2, CT\}. We include CT scans as they are used in manual delineation of the optic system. CT scans do not exhibit bias field artefacts, so we clamp the bias field parameters in our model to zero for this image type. Additionally, as CT scans have a low contrast within the brain, we can initialize the tumor-associated mean values in the same way as for normal labels. In the second combination, we only use the MR sequences, i.e., \{T1c, FLAIR, T2\}. In the last combination, we use T1c and a new combinatory sequence named FLAIR$^2$ that is designed to improve lesion detection \citep{wiggermann2016}. This image is computed by multiplying FLAIR with T2. For this image, we use the same settings in our model as for FLAIR. 
We emphasize that some of the modalities under consideration -- in particular CT and FLAIR$^2$ -- were not included in any training data available to the proposed segmentation algorithm.
We start with an overall visual inspection of the segmentations and then analyze the performance scores, followed by a more in-depth visual inspection of some of the segmentations. 

\begin{figure*}
\center
\vspace{-3mm}
\includegraphics[width=0.71\textwidth]{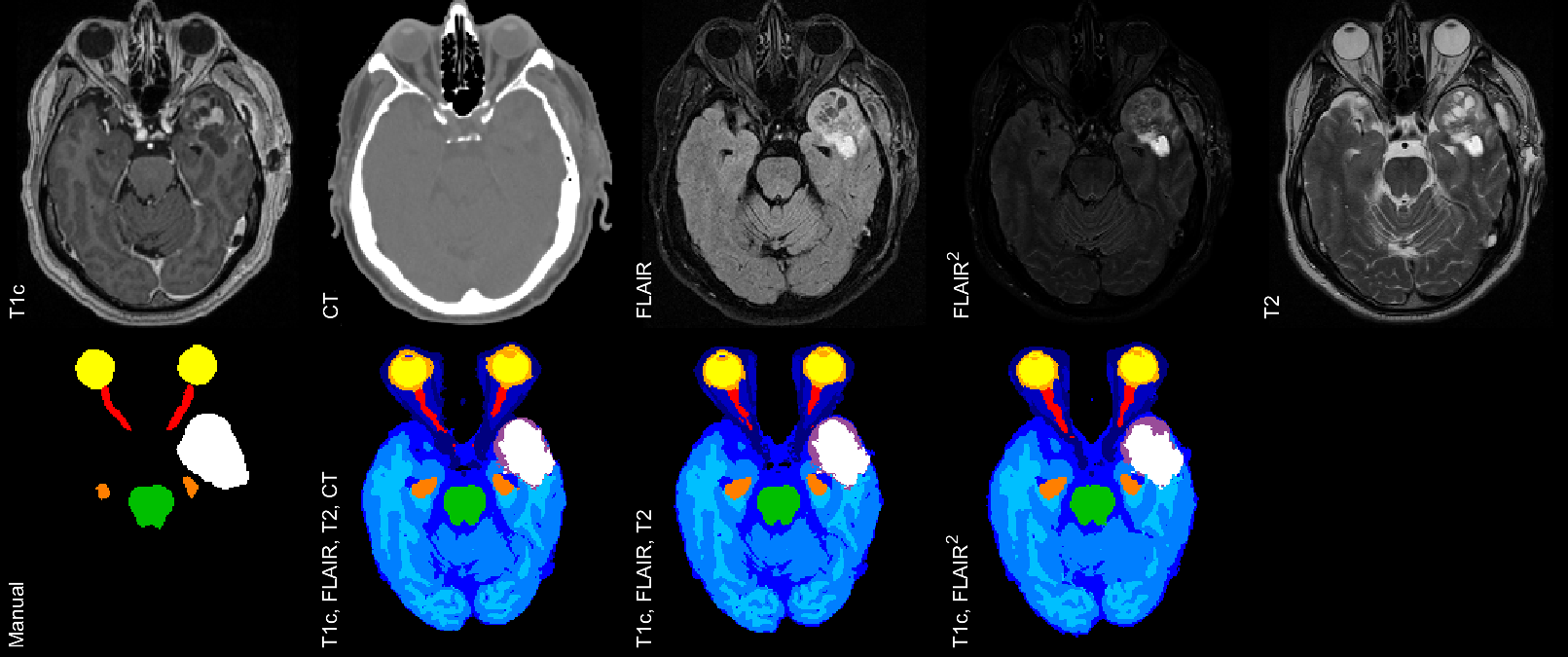}\\
\vspace{0.5mm}
\includegraphics[width=0.71\textwidth]{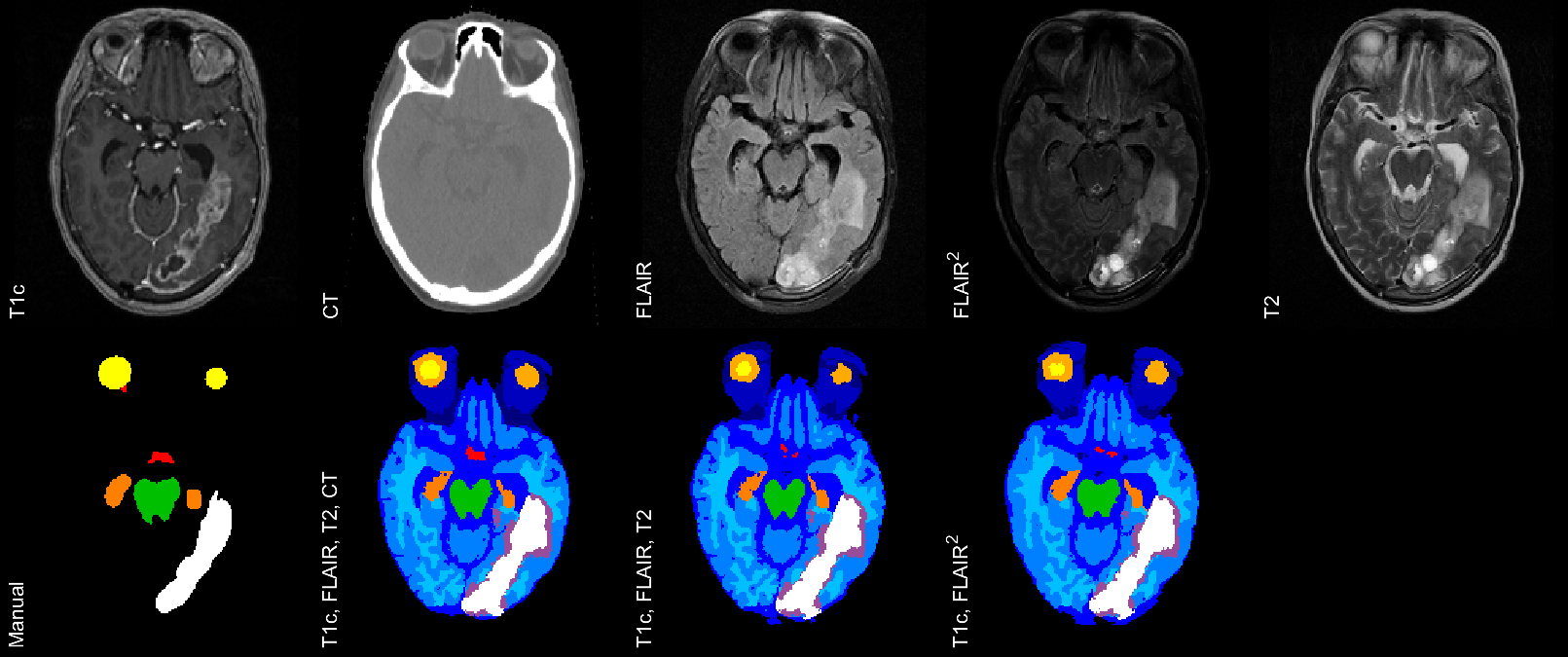}\\
\vspace{0.5mm}
\includegraphics[width=0.71\textwidth]{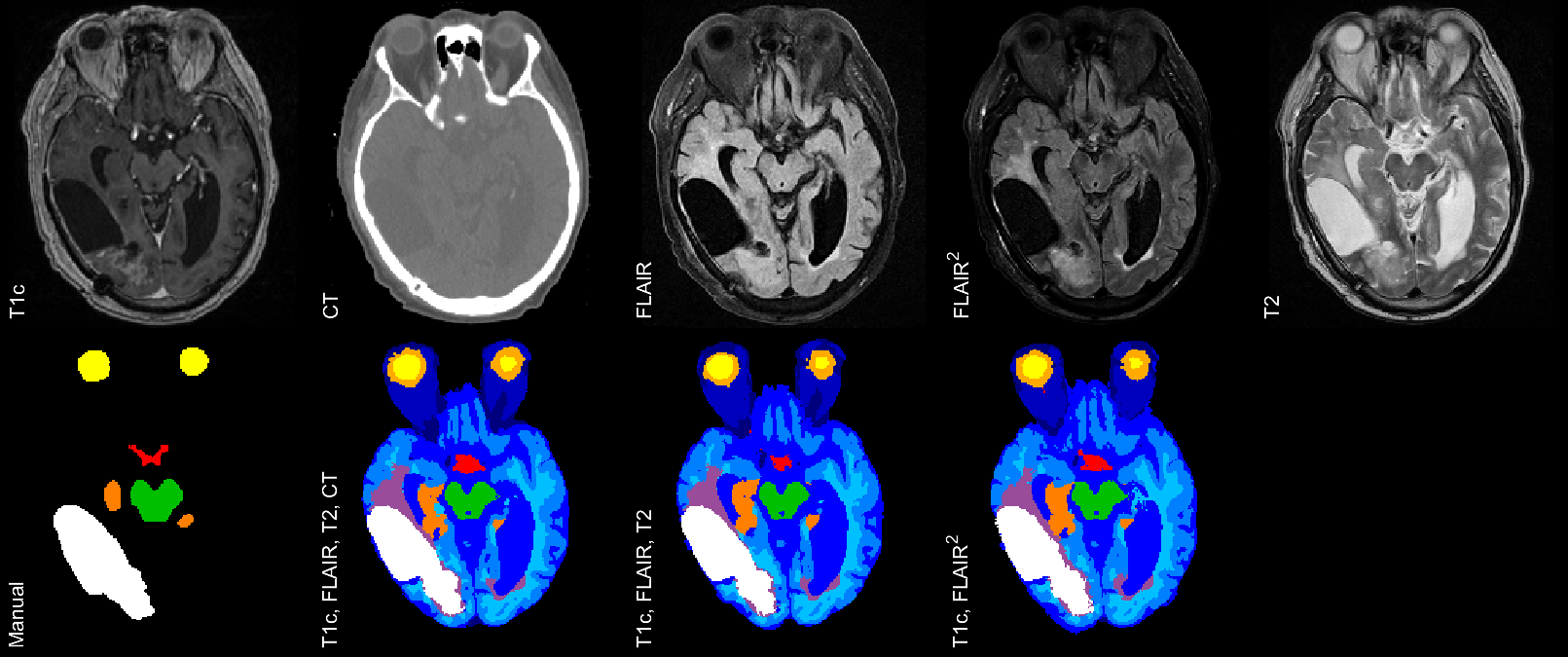}\\
\vspace{0.5mm}
\includegraphics[width=0.71\textwidth]{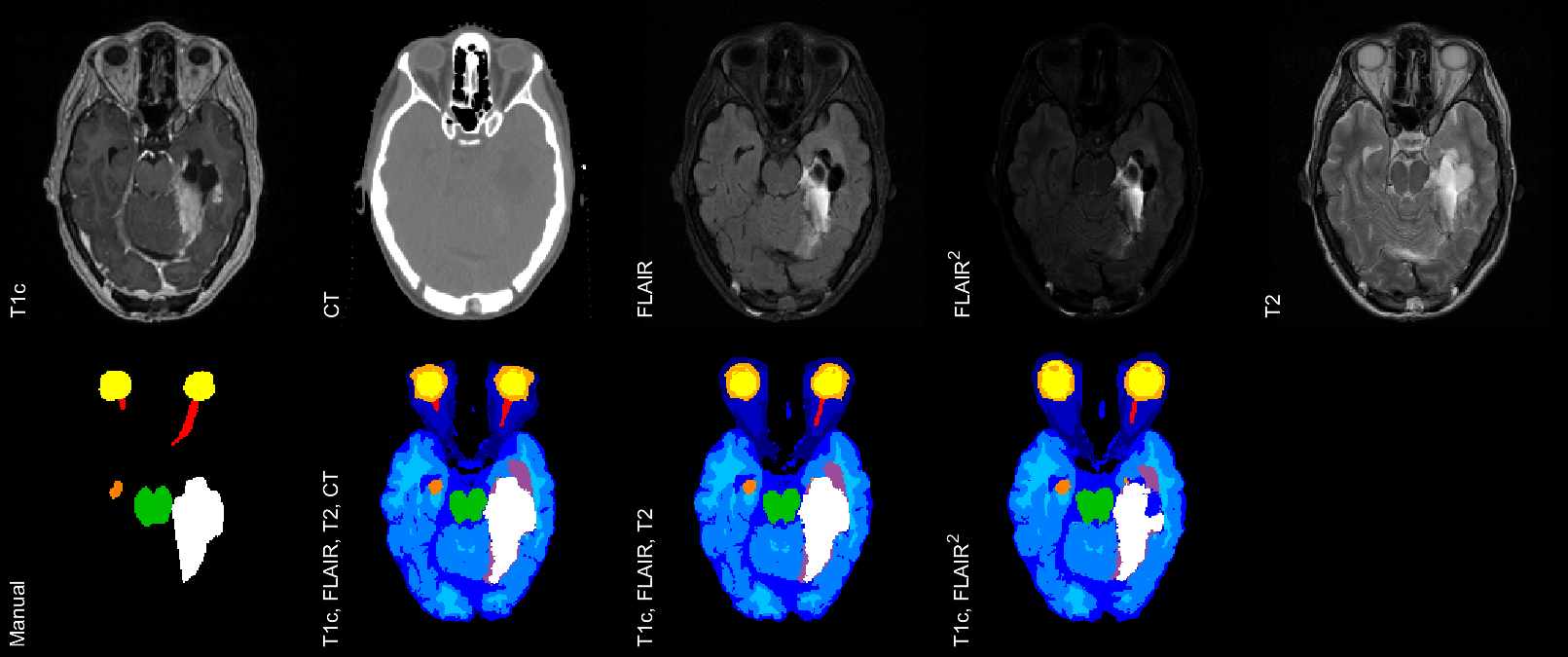}\\

\caption{Segmentations of four representative subjects in the Copenhagen dataset. 
For each subject, the top row shows slices of the data (from left to right: T1c, CT, FLAIR, FLAIR$^2$ and T2), whereas the bottom row shows, from left to right, the manual segmentation and automatic segmentations for data combinations \{T1c, FLAIR, T2, CT\}, \{T1c, FLAIR, T2\} and \{T1c, FLAIR$^2$\}. Label colors: white = TC, lilac = edema, green = BS, dark orange = HC, yellow/light orange = EB, red = ON/CH, shades of blue = other normal labels. For TC in order of appearance: Dice score: $\{0.68,0.67,0.62\}, \{0.93, 0.93, 0.91\},\{0.86,0.85,0.85\},\{0.61,0.72,0.73\}$, Hausdorff distance: $\{10,10,10\}, \{2, 3, 5\},\{7,7,6\},\{42,25,8\}$.}
\label{f:trep}
\end{figure*}

Figure \ref{f:trep} shows slices of the segmentations using the three data combinations for four representative subjects. We can see that the method in general seems to work well and consistently across all three data combinations. The atlas deforms well to fit subjects with varying shapes, and the method is capable of segmenting tumor cores of varying size, shape and intensity profile; although it underestimates the tumor size in some cases. Eyes, hippocampi and brainstem seem to be consistently well-captured, while optic nerves and chiasm are less well-captured, but better for the data combination including CT, which is because the difference in intensity between the optic nerves and surrounding tissue is larger in CT than MR. Finally, as can be noticed in the last subject, many subjects show some ambiguity in the intensity profile of the optic nerves. 

\begin{figure}
\includegraphics[width=1\columnwidth]{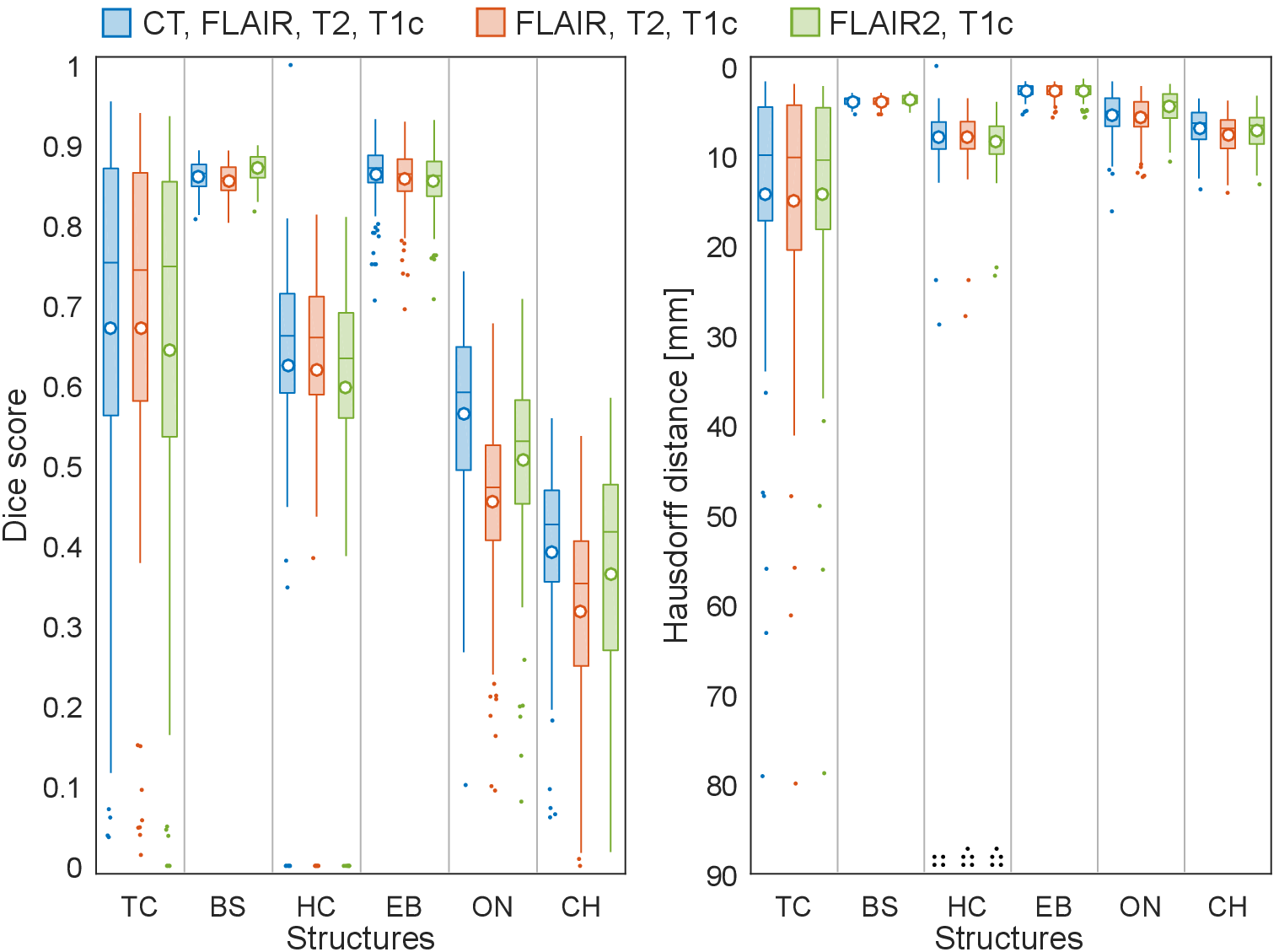}
\caption{Boxplots of Dice scores (left) and Hausdorff distances (right) for structures in the Copenhagen dataset, for three data combinations in blue, red and green, respectively. 70 subjects in total. On each box, the central line is the median, the circle is the mean and the edges of the box are the 25th and 75th percentiles. Outliers are shown as dots. Black dots at the bottom of the Hausdorff distance boxplot indicate structures for which scores could not be calculated due to missing ground truth. Note that scores for the left and right structures are included separately in the box plots for HC, EB and ON.}
\label{f:cscores}
\end{figure}

Figure \ref{f:cscores} shows box plots of the Dice scores and Hausdorff distances for the three data combinations, with the following structures: tumor core (TC), brainstem (BS), hippocampi (HC), eyes (EB), optic nerves (ON), and chiasm (CH). The left and right structures are included as separate scores in the plots for hippocampi, eyes, and optic nerves. As can be seen, the method readily adapts to the various included and excluded images in the three data combinations without the need for adjustment. The scores are consistent across the three data combinations for all regions except optic nerve and chiasm. The average Dice scores for tumor core are fair, but the range of scores is large. However, this is consistent with the state of the art in brain tumor segmentation, as will be shown in Section \ref{sec:brats}. Furthermore, this dataset includes a number of difficult subjects with large resections, small and thin contrast-enhanced tumor regions in T1c and small bright tumor regions in FLAIR. 

The Dice scores for brainstem are high and consistent across the subjects and comparable to the ones obtained with the healthy whole-brain segmentation method that our method is based on \citep{puonti2016}. Furthermore, the Hausdorff distances are low and consistent as well. For eyes, the Dice scores are generally high, except for a few outliers that were affected by a very thin outer eye wall, and the Hausdorff distances are generally low, indicating a good performance.  Hippocampi, on the other hand, have a range of generally lower Dice scores than in \citep{puonti2016}. 
Their Hausdorff distances are also fairly large. In the majority of the outliers, the method has segmented the hippocampus near to the tumor border while the manual segmentations either lack that hippocampus or have undersegmented it. Finally, the Dice scores for optic nerves and chiasm are generally low and with a large range. These structures are very small and thin, which significantly affects this metric. The Hausdorff distances for these structures are reasonably low however, which indicates that the manual and automatic segmentations are in fact fairly close. The Dice scores for the data combination including CT are higher, due to the better contrast in CT between the optic nerve and surrounding structures.
 
\begin{figure*}
\vspace{-10mm}
\centering
\begin{tabular}{>{\centering}m{0.55\textwidth} m{0.33\textwidth}}
\includegraphics[width=0.55\textwidth]{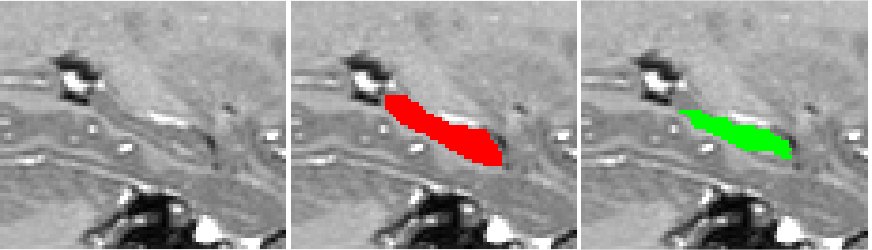} & \includegraphics[width=0.33\textwidth]{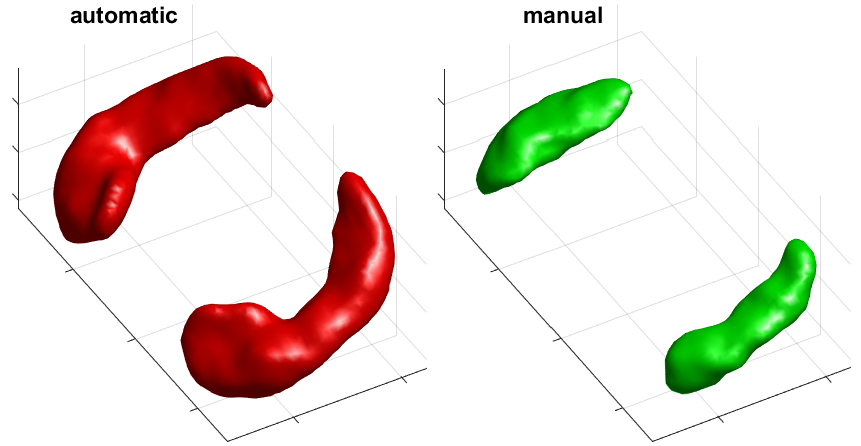} \\
\includegraphics[width=0.55\textwidth]{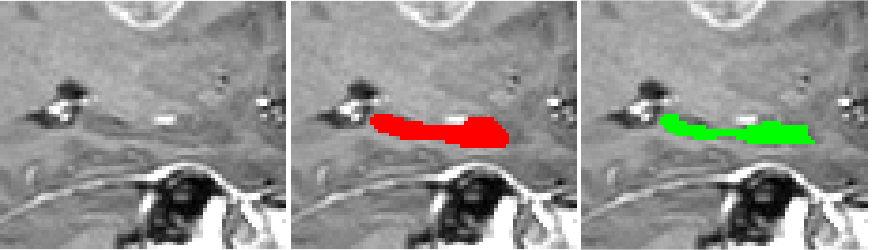} & \includegraphics[width=0.33\textwidth]{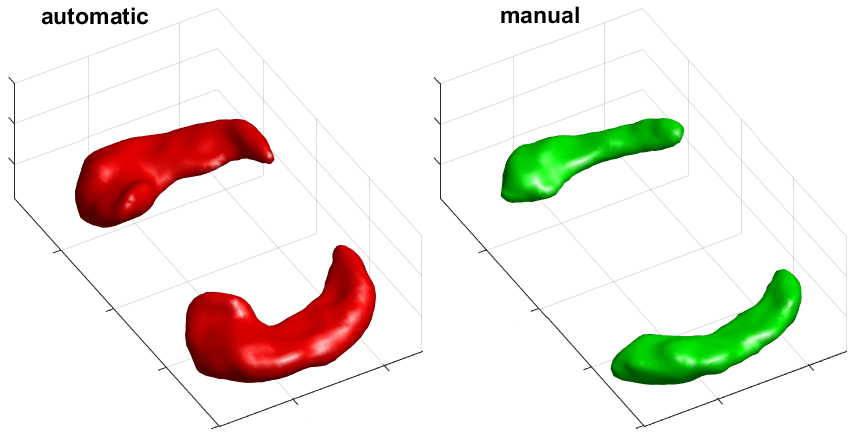} \\
\end{tabular}
\caption{Hippocampi on two representative subjects in the Copenhagen dataset. Automatic segmentations (for \{T1c, CT, FLAIR, T2\}) in red and manual segmentations in green. Slice of segmentation overlaid on the T1-weighted scan and 3D surface plot of full structure. For left and right hippocampus: Dice score: $\{0.54,0.58\}$,$\{0.63,0.67\}$; Hausdorff distance: $\{13,10\}$, $\{8,7\}$.}
\label{f:hipp}
\end{figure*}

Figure \ref{f:hipp} shows sagittal slices of two representative segmentations of hippocampi, together with surface plots of the manual and automatic segmentation (for \{T1c, CT, FLAIR, T2\}). In both cases, the automatic segmentations are larger and seem to capture the hippocampi somewhat better than the manual segmentations. As can be seen in the surface plots, the manual segmentations are not very consistent with each other. The head and subiculum of the hippocampi are also excluded, due to a difference in segmentation protocol compared to the healthy segmentations used to build the method's atlas. To a large extent, this explains the fairly low and inconsistent Dice scores. Another reason for the lower Dice scores compared to \citep{puonti2016} could be the large slice thickness in FLAIR and T2, which introduces large partial volume effects.  

\begin{figure*}
\vspace{-3mm}
\centering
\begin{tabular}{>{\centering}m{0.57\textwidth} m{0.26\textwidth}}
\includegraphics[width=0.57\textwidth]{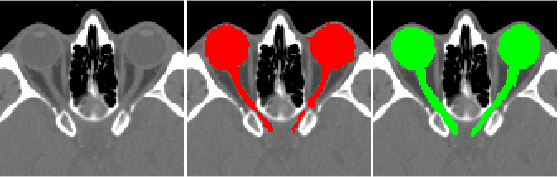}&
\includegraphics[width=0.26\textwidth]{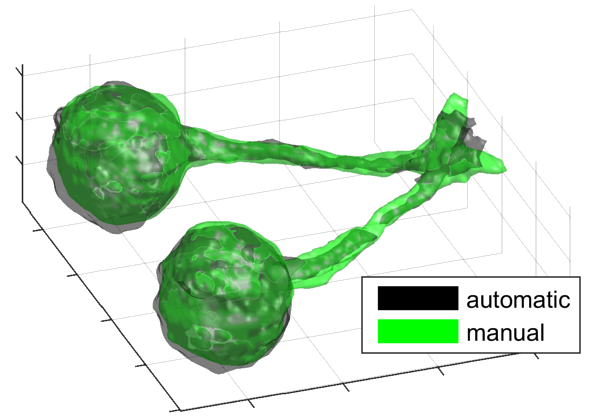}\\
\includegraphics[width=0.57\textwidth]{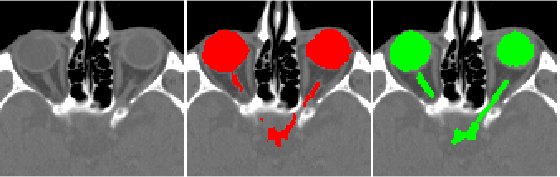}&
\includegraphics[width=0.25\textwidth]{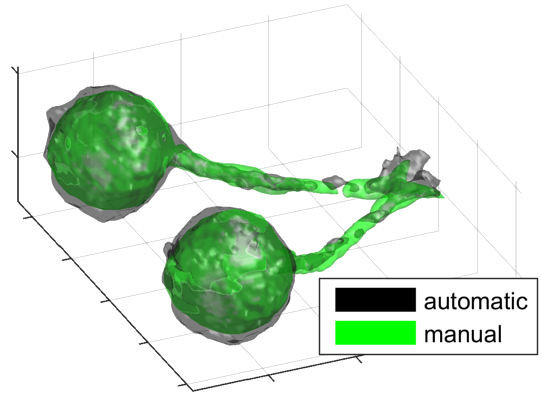}\\
\end{tabular}
\caption{Optic system on two representative subjects in the Copenhagen dataset. Automatic segmentations (for \{T1c, CT, FLAIR, T2\}) in red and manual segmentations in green. Slice of segmentation overlaid on the CT scan and 3D surface plot of full structure. For right and left eye; right and left optic nerve; and chiasm: Dice score: $\{0.91,0.89\}$, $\{0.91,0.87\}$, $\{0.67,0.67\}$, $\{0.48,0.55\}$ and $\{0.49,0.44\}$; Hausdorff distance: $ \{2,2\}$, $\{2,2\}$, $\{4,4\}$, $\{4,6\}$ and $\{4,6\}$.}
\label{f:eye}
\end{figure*}

Figure \ref{f:eye} shows slices of two representative segmentations of the optic system (including eyes, optic nerves and chiasm), together with surface plots of the manual and automatic segmentation (for \{T1c, CT, FLAIR, T2\}). The method captures the eyes well, although in some cases the wall of the eye is slightly oversegmented. By visual inspection, we found that the method has some difficulties when a subject has the eye lids open, as the solid wall between eye and air becomes very thin. Furthermore, when guided by CT, the method captures the optic nerve (the thin nerve going from an eye in one end to the chiasm in the other end) reasonably well. However, the method has problems in the region where the nerve goes through the skull, as the nerve is especially thin in this region. Because the nerve is thin, the method is also sensitive to intensity ambiguities in the data, such as artifacts or movement of the optic nerve between image acquisitions. In general, the method finds the location of chiasm, but because this structure is so small, the segmentation is to an even larger extent affected by partial volume effects and intensity ambiguities. Finally, the manual segmentations are quite variable in where the borders are placed between the optic nerves and chiasm, as well as between chiasm and the optic tracts (the continuation of the optic system into the brain).

\begin{figure*}
\center
\vspace{-3mm}
\includegraphics[width=0.7\textwidth]{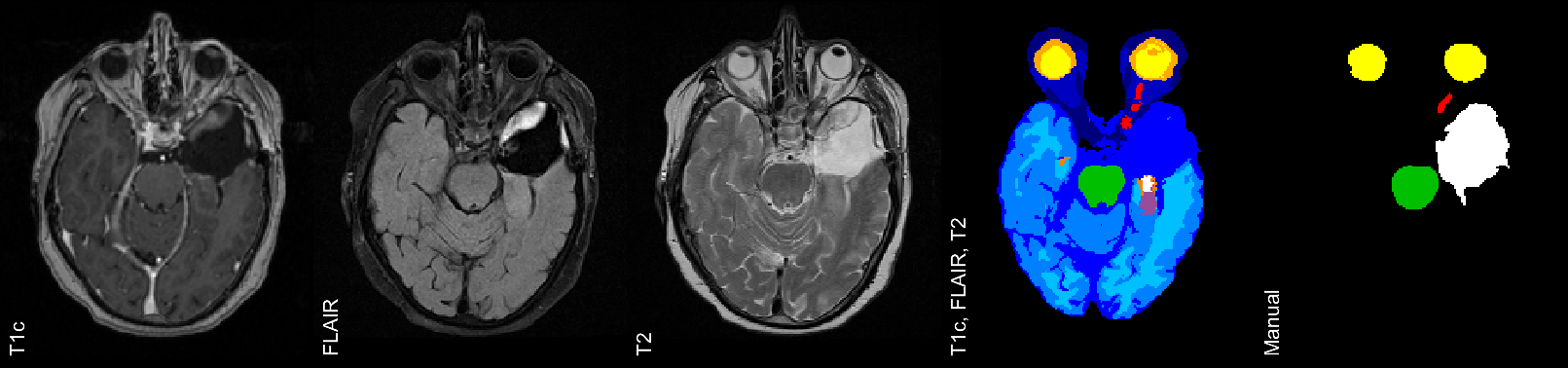}\\
\vspace{0.5mm}
\includegraphics[width=0.7\textwidth]{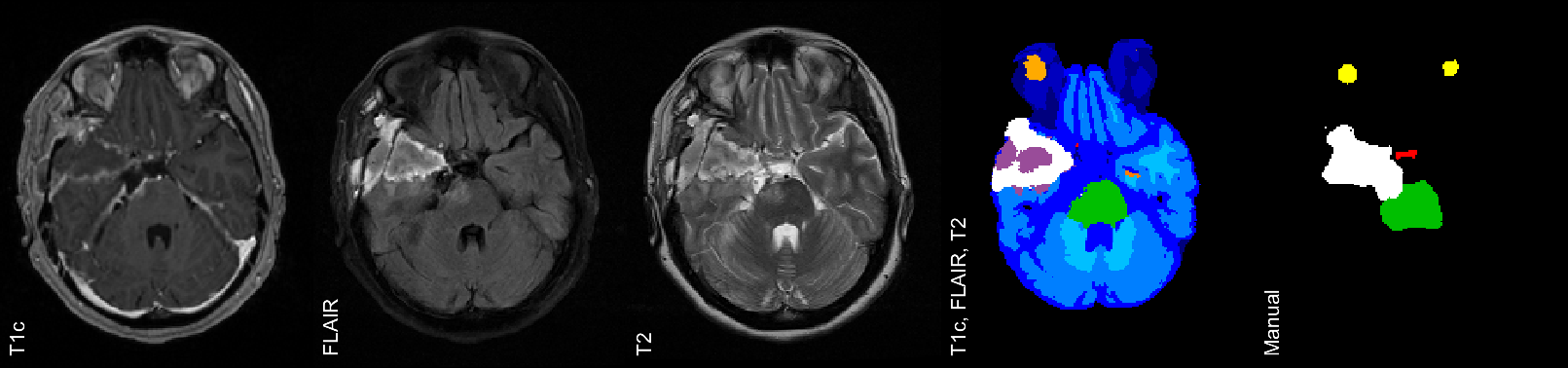}\\
\caption{Two problematic tumor core segmentations in the Copenhagen dataset. Data slices shown together with automatic segmentation (for \{T1c, FLAIR, T2\}) and manual segmentation. For tumor core: Dice score: $\{0.04, 0.45\}$, Hausdorff distance: $\{41, 28\}$.}
\label{f:failed}
\end{figure*}

Figure \ref{f:failed} shows slices of two problematic tumor core segmentations (for data combination \{T1c, FLAIR, T2\}) that are representative of cases when the method struggles. The first case includes a very large resection at the border of the brain, which the method has difficulty to adapt to for three main reasons: (1) resectioned tumor regions close to the border of the brain can be interpreted as CSF by the method; (2) the method relies on the contrast-enhanced tumor region, which in this case is thin and with weak contrast-enhancement; (3) the method also relies on a bright tumor region in FLAIR, which in this case is small and only slightly brighter than surrounding tissue. In the second case, the method struggles to fill in the inner part of the tumor core. This is an issue in a few cases where the intensity profile of the inner part of the core is similar to that of edema or healthy tissues. 


\subsubsection{Dosimetric evaluation} 
To estimate whether the use of automatic, rather than manual, segmentations introduces any differences in metrics typically reviewed when planning a radiation therapy session, we conduct an additional dosimetric evaluation of our results.

\begin{figure}[ht]
\center
\includegraphics[width=0.5\columnwidth]{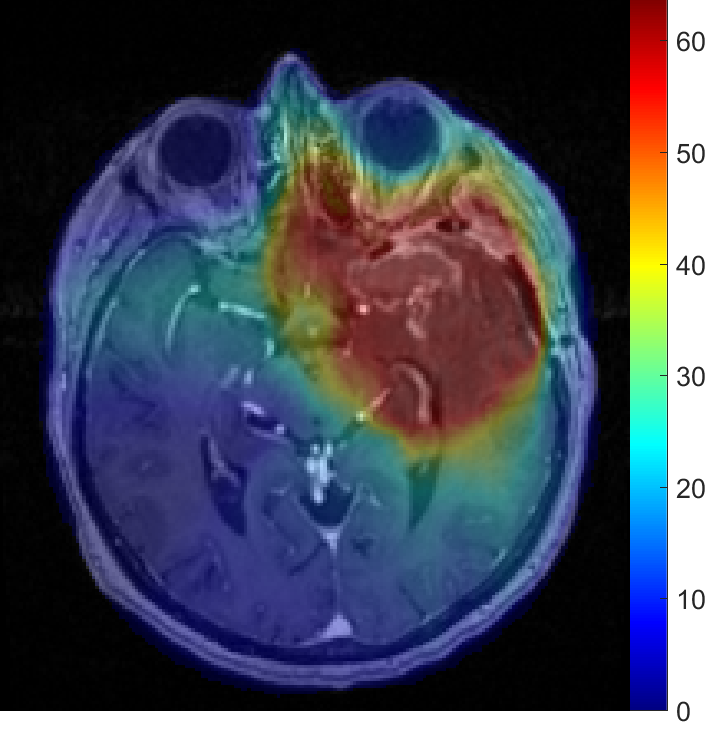}
\caption{A radiation dose plan overlaid on a T1c image slice for a representative subject. The dose is measured in Gy.}
\label{f:dplan}
\end{figure}

During radiation therapy planning, the segmentations of tumor core (clinically defined as GTV) and OARs are used to optimize a radiation dose distribution that will be used during treatment. Figure \ref{f:dplan} shows an example of such a radiation dose plan. Note that, to form the target to be irradiated, a margin is added around the tumor core to cover likely subclinical spread of tumor cells, which is defined as the clinical target volume (CTV). Finally, a margin stemming from any geometrical uncertainties adhering to the treatment planning and radiation delivery is added, and this volume is defined as the planning target volume (PTV). During the treatment planning process, each OAR and target structure (usually only the PTV) is given a dose-volume objective and a priority that varies with the clinical relevance. A more detailed explanation of the dose plan optimization is given in \citep{munck2011}. 

\begin{figure}
\center
\includegraphics[width=0.9\columnwidth]{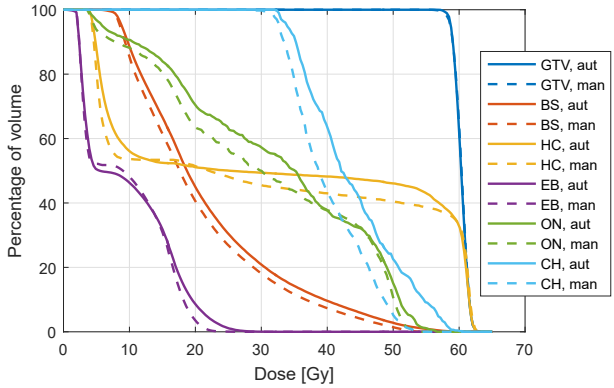}
\caption{Dose volume histogram (DVH) of several structures for the representative subject in Figure \ref{f:dplan}, i.e., tumor core (GTV), brainstem (BS), hippocampi (HC), eyes (EB), optic nerves (ON), and chiasm (CH). Solid lines and broken lines correspond to automatic and manual segmentations, respectively.}
\label{f:dvh}
\end{figure}

To assess the delivered dose to different structures, cumulative dose-volume histograms (DVHs) are often used. Each bin in a DVH represents a certain dose and shows the volume percentage of a structure that receives at least that dose. Figure \ref{f:dvh} shows the DVHs of all relevant structures for the example in Figure \ref{f:dplan}, i.e., tumor core (GTV), brainstem (BS), hippocampi (HC), eyes (EB), optic nerves (ON), and chiasm (CH). We show DVHs for both the manual and the automatic segmentations for the data combination \{T1c, CT, FLAIR, T2\}. All histograms are computed using the dose plan calculated for the planned treatment, i.e., using the manual segmentations. The wide margin added around the tumor core means that the hippocampus in the same hemisphere is frequently located almost completely inside the tumor target. This is the case for the example we show,
which is why almost half of the hippocampi volume is irradiated as much as the tumor core,
as can be seen in Figure \ref{f:dvh}. The maximum accepted dose to the optic chiasm and optic nerves during the treatment planning phase is generally 54 Gy, though small volumes may exceed that dose occasionally. Using the automatic segmentation of the optic chiasm, the radiation dose maximum is somewhat above 54 Gy, suggesting some clinically relevant disagreement between the manual and automatic chiasm segmentations. 

\begin{figure}[ht]
\includegraphics[width=1\columnwidth]{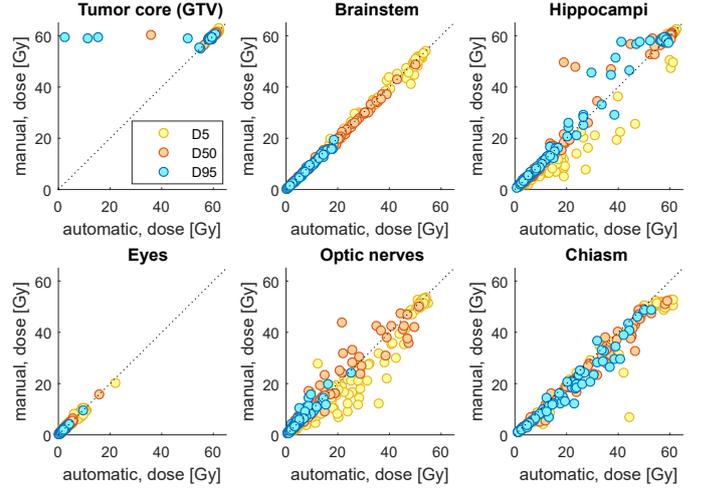}
\caption{Summary statistics of DVH results for all subjects and structures, showing 5\% volume (D5), 50\% volume (D50), and 95\% volume (D95), for manual versus automatic segmentations. Note that left and right hippocampus, eye and optic nerve are included as separate points in their respective plots.}
\label{f:dvhAllSub}
\end{figure}

To ease the comparison of the DVH results of the automatic and manual segmentations for all subjects, we summarize them as in \citep{conson2014} by using three points in the histograms. To cover a large part of the cumulative histograms, 
we use the dose at 5\% of volume (D5), 50\% of volume (D50), and 95\% of volume (D95). Figure \ref{f:dvhAllSub} shows the summarized results for all structures, with values for the manual segmentations plotted against values for the automatic segmentations. In the plots, the closer a point is to the diagonal line, the closer the results of the manual and automatic segmentations are. For tumor core, most points are very close to the line, which is unsurprising considering the wide margin added around tumor. The four D95 outliers belong to subjects where small regions in the brain were erroneously segmented as tumor core by our method, for some cases because of co-occurring pathologies. The results for the organs-at-risk largely confirm the findings using Dice scores and Hausdorff distances. Brainstem and eyes are delineated in close agreement, and the issue with oversegmentation when the outer eye wall is very thin does not affect the dosimetric measure, because that region will always be far away from tumor. The results for hippocampi are varying for subjects where a hippocampus is on the border of the tumor target, mainly due to the difference in protocol between the manual and automatic segmentations. Furthermore, the results for optic nerves vary widely for a few subjects. However, at the maximum dose target of 54 Gy the results of the manual and automatic segmentations match fairly well. For the optic chiasm, on the other hand, some results for the automatic segmentations are significantly beyond its dose objective of maximum 54 Gy. This suggests that significant differences to treatments could be expected if the automatic segmentation of this structure would be used instead of the manual segmentation when optimizing the radiation dose plan. 


\subsubsection{Comparing our tumor prior to first-order MRFs} 
To demonstrate the benefits of modeling high-order interactions with the cRBM-based tumor prior, we will contrast it to a tumor prior based on more traditional first-order MRFs. As mentioned before, first-order MRFs only have pairwise clique potentials, compared to the potentials in cRBMs that are defined over groups of voxels as large as the size of the convolutional filters. The inference of the model is kept exactly the same except for the tumor prior in Algorithm \ref{a1}: there are no hidden units to sample and 
therefore
the labels in a voxel $i$ are sampled from
\begin{multline*}
  p_i( l_i,z_i,y_i | \mathbf{d}_i, \boldsymbol \theta, \hat{\boldsymbol \eta})
  \propto
  \\
  \quad
  \begin{array}{l}
  p_i( \mathbf{d}_i | l_i,z_i,y_i, \boldsymbol \theta )
  \pi_i( l_i )
  \exp\left[ 
    -\beta^z \sum_{j \in \mathfrak{N}_i} |z_i - z_j| 
  \right]  
  \\
  \qquad
  \exp\left[ 
    -\beta^y \sum_{j \in \mathfrak{N}_i} |y_i - y_j|
  \right]
  \exp\left[ 
    -f(l_i,z_i,y_i)
  \right]
  ,
  \end{array}
\end{multline*}
where $\mathfrak{N}_i$ is the set of 26 voxels that form neighboring pairs with voxel $i$.

To find suitable values for the user-tunable hyperparameters $\beta_z$ and $\beta_y$, we performed a grid search with steps of 0.5 using the same 30 manually segmented BRATS training subjects that we used for training the cRBMs (see Section \ref{s:rbm}). For each hyperparameter combination, we segmented the subjects using the method with first-order MRFs. By comparing the average performance (using Dice scores and Hausdorff distances) we found the combination $\{\beta_z = 4, \beta_y = 1\}$ to have the best overall performance.  
With these optimized hyperparameter values,
we compare the tumor core segmentation performance on the data combination \{T1c, FLAIR, T2\} when using the two different priors. The average and median Dice score for the cRBM-based method is 0.67 and 0.74 respectively, compared to 0.58 and 0.57 when using the first-order MRFs described here. Furthermore, the average and median Hausdorff distance for the cRBM-based method is 14 mm and 10 mm respectively, compared to 23 mm and 17 mm when using first-order MRFs. This demonstrates the benefit of modeling high-order interactions among voxels.

\subsection{Results for 2015 BRATS test dataset}
\label{sec:brats}
To further evaluate our method's performance on segmenting tumors and compare it to that of other methods, we use the test dataset of the 2015 BRATS challenge. We participated in this challenge and were among the top-performing methods out of a total of 12 methods. 
This dataset includes non-enhanced T1 scans, which the dataset in Section \ref{sec:clin} lacks, and data with varying magnetic field strength and resolution from several imaging centers. 
The dataset is skull-stripped, so we merge all non-brain labels used in our method into the background label. We stress that we did not need to change anything else in our method.

The dataset is publicly available at the virtual skeleton online platform \citep{virtualskeleton}. It consists of 53 patients with varying high- and low-grade gliomas. Some of the patients have had tumor resections. The included MR contrasts are T2-weighted FLAIR (2D acquisition), T2-weighted (2D acquisition), T1-weighted (2D acquisition) and T1-weighted with contrast enhancement (T1c, 3D-acquisition). All data were resampled to 1 mm isotropic resolution, aligned to the same anatomical template and skull-stripped by the challenge organizers. The dataset includes manual annotations of four tumor regions, which are not publicly available. Instead, the performance of a method can be evaluated by uploading segmentations to the online platform. On the online platform and during the challenge, scores are reported on enhanced core, core (which includes enhanced core and other core regions), and whole tumor (which includes core and edema). Note that there have been two challenges since the 2015 challenge, but without any publicly released test data as of yet. In addition, only preoperative scans are included in the most recent BRATS challenge (2017). Thus, it is less relevant in a radiation therapy context, as radiation therapy patients frequently have had resections.

\begin{figure}[H]
\center
\includegraphics[width=0.28\columnwidth]{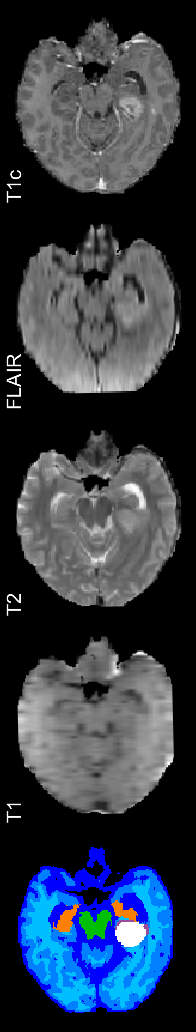}
\includegraphics[width=0.28\columnwidth]{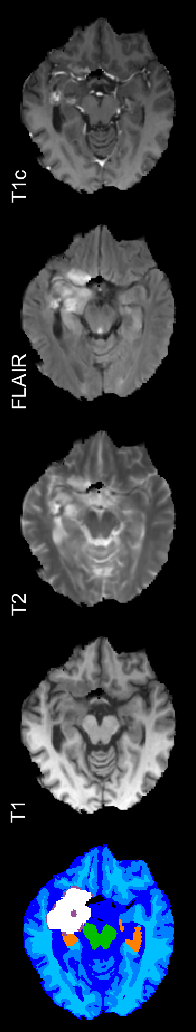}
\includegraphics[width=0.28\columnwidth]{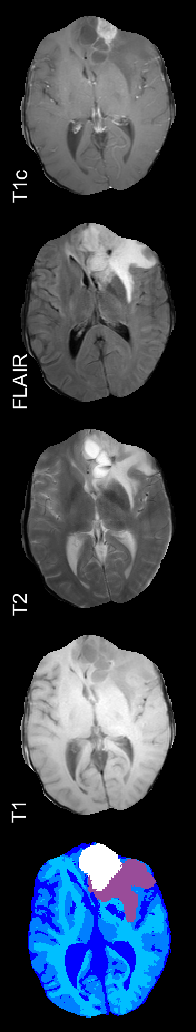}
\caption{Three representative segmentations in the BRATS test dataset. Slices of T1c, FLAIR, T2, T1, and automatic segmentation. Label colors: white = TC, lilac = edema, green = BS, dark orange = HC, shades of blue = other brain tissues. Note that the images are skull-stripped by the BRATS challenge organizers.} 
\label{f:trepBRATS}
\end{figure}

Figure \ref{f:trepBRATS} shows slices of three representative segmentations with: T1c, FLAIR, T2 and T1, and the segmentation by our method as presented in this paper. Note that the manual segmentations compared against are not publicly available. We can see that the atlas deforms well to the subjects, and brainstem and hippocampi are well-captured. Furthermore, our method can segment brain tumors with large variations in size, location and appearance. Also note the low resolution and image quality in some of the images. 

For the purpose of comparing against the manual segmentations,
we focus on the core region, as this corresponds to the GTV used in radiation therapy. We compare the performance of our method to that of three other top-performing tumor segmentation methods that also participated in the 2015 BRATS challenge.

\textit{1) GLISTRboost \citep{bakas2016}:} This semi-automated method is based on a modified version of the generative atlas-based method GLISTR \citep{kwon2014,gooya2012}, which uses a tumor growth model. The method requires manual input of a seed-point for each tumor center and a radius of the extent of the tumor. To increase the segmentation performance, the method is extended with a discriminative post-processing step using a gradient boosting multi-label classification scheme followed by a patient-wise refinement step. 

\textit{2) Grade-specific CNNs \citep{pereira2016}:} This semi-automated method uses a discriminative 2D Convolutional Neural Network (CNN) approach. The method takes advantage of the fact that high- and low-grade tumors exhibit differences in intensity and spatial distribution. To do this, it uses two CNNs: one trained on high-grade tumors and one trained on low-grade tumors. The CNN to use for a specific subject is then chosen manually based on visual assessment, which is the only manual step in the method. 

\textit{3) Two-way CNN \citep{havaei2016}:} This fully automated method uses a similar discriminative 2D CNN approach to the previous method. The method forms a cascaded architecture with two parts, where the voxel-wise label predictions from the first part are added as additional input to the second part. Each part has two pathways, where intensity features are automatically learned: one learning local details of tumor appearance and one learning larger contexts. 

\begin{figure}[H]
\center
\includegraphics[width=0.9\columnwidth]{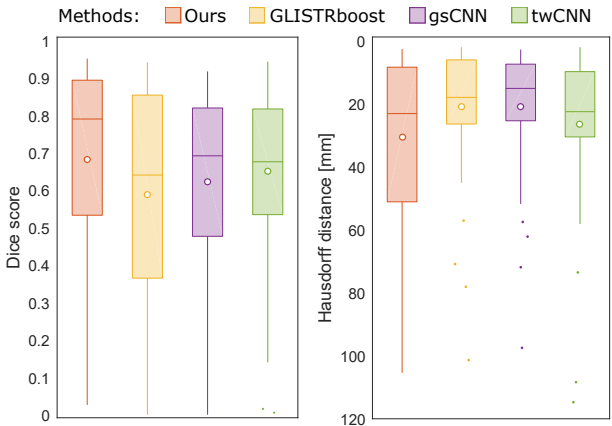}
\caption{Box plots of Dice scores and Hausdorff distances for tumor core on the BRATS 2015 test dataset. 53 subjects in total. Scores are as reported in the challenge. On each box, the central line is the median, the circle is the mean and the edges of the box are the 25th and 75th percentiles. Outliers are shown as dots.}
\label{f:bscores}
\end{figure}

Figure \ref{f:bscores} shows box plots of the Dice scores and Hausdorff distances for tumor core. We show scores for our method and the three benchmark methods as reported at the challenge. The scores for our method are for the version we participated with in the challenge, as presented in \citep{Agn2016}. The main difference, compared to the current version, is the use of an affinely registered atlas, instead of the mesh-based deformable atlas presented in this paper to enable a detailed segmentation of normal head structures. This, however, does not significantly affect the tumor segmentation; we also segmented the dataset with our current version and obtained similar Dice scores from the online platform, with just a 4 \% increase in the average Dice score. As seen in the figure, the range of Dice scores is similar to our results in Section \ref{sec:clin} (Figure \ref{f:cscores}), which shows that our method readily adapts to the included non-enhanced T1 scans and data from different imaging centers. Comparing to the other benchmark methods, our method performs significantly better on tumor core when considering Dice scores. The range of values are large for all methods, illustrating the difficulty of segmenting tumors. This dataset includes a number of subjects with large resections and a wide variety of tumors, e.g., low-grade tumors that have been shown to be difficult to segment in \citep{menze2015}. The Hausdorff distances for our method are somewhat worse than for the other methods, which could be explained by a better capability of their methods to remove small erroneous tumor clusters, e.g., because of the deep architecture in a CNN. The Hausdorff distances for our method are also worse for this dataset than for the dataset in Section \ref{sec:clin} (cf. Figure \ref{f:cscores}), which is explained by the generally lower resolution and image quality.

\subsection{Results for London dataset}
\label{sec:dir}
As a final experiment, we investigate the ability of our method to adapt to 
yet a different set of acquired images using the London dataset. 
In contrast to the other datasets, this one completely lacks T1-weighted images and includes a new MR sequence: double inversion recovery (DIR). The data set consists of seven patients with varying low- and high-grade gliomas, which were scanned with a Siemens Trio 3T scanner at the National Hospital for Neurology and Neurosurgery, UCLH NHS Foundation Trust, London, as part of a registered clinical audit. The following MR images were acquired with 1 mm isotropic resolution: T2-weighted (3D acquisition) and T2-weighted DIR (3D-acquisition). We use exactly the same settings in our method for the DIR images as we would for FLAIR, without any changes. As no manual segmentation has been performed on this dataset, we only 
perform a qualitative analysis
of the results.

\begin{figure}[h]
\center
\includegraphics[width=0.9\columnwidth]{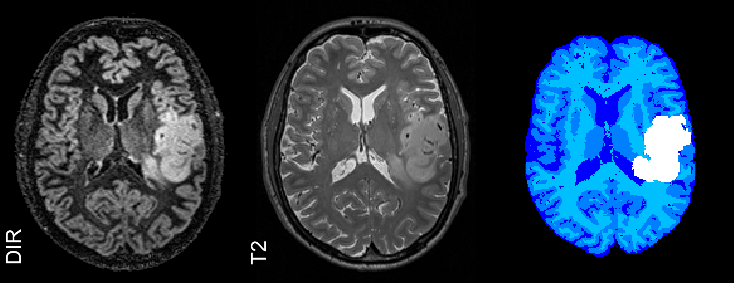}\\
\vspace{0.5mm}
\includegraphics[width=0.9\columnwidth]{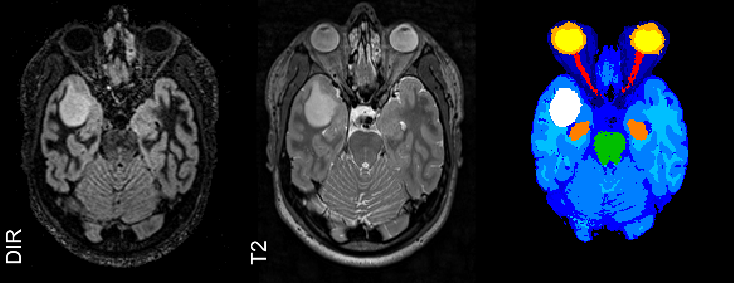}\\
\vspace{0.5mm}
\includegraphics[width=0.9\columnwidth]{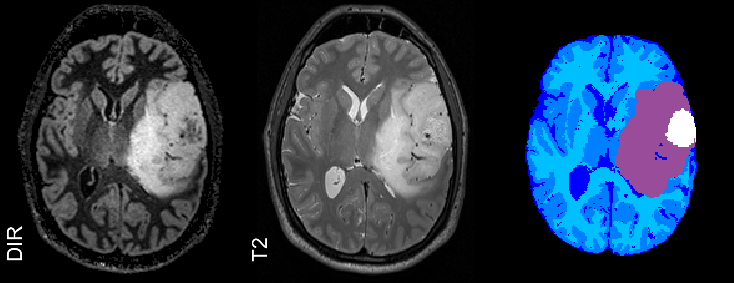}\\
\caption{Three representative segmentations in the London dataset. Slices of DIR, T2 and automatic segmentation.}
\label{f:dir}
\end{figure}

Figure \ref{f:dir} shows slices for three representative subjects with DIR, T2 and the method's segmentation. As seen in the Figure, our method can easily segment datasets that lack T1-weighted images and include a DIR image instead of FLAIR without any changes to the method. Visual inspection of all seven segmentations revealed no significant deviations from other results presented in this paper. 


\section{Discussion and conclusion}
\label{sec:conclusion}
In this paper, we have presented a 
generative method for simultaneous segmentation of brain tumors and an extensive set of 
organs-at-risk (OARs) 
applicable to radiation therapy planning for glioblastomas. To the best of our knowledge, this is the first time a segmentation method has been presented that encompasses both brain tumors and OARs within the same modeling framework. The method combines a previously validated atlas-based model for detailed segmentation of normal brain structures with a model for brain tumor segmentation based on convolutional restricted Boltzmann machines (cRBMs). In contrast to generative lesion shape models proposed in the past, 
cRBMs are capable of modeling long-range spatial interactions among tumor voxels.
Furthermore, 
by completely separating
the modeling of anatomy from the modeling of image intensities, 
the method is able
to adapt to heterogeneous data that differs 
substantially
from any available training data,
including unseen (e.g., CT or FLAIR$^2$) or missing (e.g., T1) contrasts.

Although the method we propose is demonstrated to be 
applicable across data with various image contrast properties without 
retraining,
it does rely on contrast-specific settings to constrain and to initialize 
tumor-specific appearance parameters,
especially in 
the MR-sequences FLAIR and T1c 
(see Table \ref{t:const} and Table \ref{t:distances}, respectively). 
We found that this was necessary to guide the model to the correct intensities for tumor 
in these sequences, which are typically acquired for brain tumor imaging. 
The results demonstrate that the same settings work robustly across FLAIR and T1c images acquired with a variety of scanners and imaging protocols, and even when FLAIR is replaced with FLAIR$^2$ or DIR.
In data where FLAIR and/or T1c is entirely missing, 
however, 
the method may need to be adjusted 
by modifying the corresponding lines in Tables \ref{t:const} and \ref{t:distances}. 

Our experiments show that the method's performance in segmenting tumors 
is comparable
to other state-of-the-art methods in brain tumor segmentation, while also being capable of segmenting the OARs hippocampi, brainstem, eyes, optic nerves and optic chiasm. We quantitatively evaluated the OAR segmentations in 70 patients with manual segmentations used when planning a radiation therapy session. The evaluation showed a generally good performance in segmenting hippocampi (HC), brainstem (BS) and eyes (EB); but lower performance in segmenting the very small structures optic nerves (ON) and chiasm (CH). 
The overall performance of our method
(average Dice scores for BS: 0.86, EB: 0.86, ON: 0.56, CH: 0.39 when using the image combination \{CT, T1c, FLAIR,T2\})
is comparable
to the human inter-rater variability reported in \citep{deelay2011}, where eight experts segmented OARs in 20 high-grade glioma patients,
with average Dice scores BS: 0.83, EB: 0.84, ON: 0.50, CH: 0.39. 
It is clear that the Dice scores for optic nerves and chiasm can be low even for experts. 
Nevertheless, the dosimetric evaluation and visual inspection of our automated segmentation of these structures point to the need for further research to obtain better results.  An improvement could possibly be achieved by incorporating dedicated geometrical information in the prior, e.g., about the tubular structure of the optic system which was successfully used in \citep{Noble2011}.

Using manual segmentations from radiation therapy planning as ground truth complicates our findings to some degree, as these segmentations themselves might be suboptimal with large inter-rater variability. Different clinics might also use differing delineation protocols. In our experiments, the Dice score for hippocampi was significantly affected by differing delineation protocols between the experts at the clinic and the expert segmentations used to train the atlas in our method. The manual segmentations at the clinic were also found to be 
of variable quality
in regions where the segmented structures have a similar intensity profile to neighboring structures -- such as the chiasm and brainstem compared to neighboring white matter structures. Additionally, structures far away from a tumor are sometimes not carefully delineated because they will not significantly affect the radiation therapy plan anyway. 

The original segmentations we used to train our atlas for normal brain structures include dozens of segmented structures. The method could handle any of these structures by simply retraining the atlas 
on segmentations in which these structures have not been merged into global catch-all labels as we did in the current paper.
This may be helpful if additional OARs need to be segmented or for automating CTV decisions
based on anatomical context \citep{unkelbach2014radiotherapy}.
A detailed whole-brain segmentation can also be useful for training outcome prediction models, e.g., to study the effect of the radiation received by various structures on cognition \citep{conson2014}.

The tumor segmentation performance of our method 
is comparable to that of
other brain tumor segmentation methods, but should still be improved for clinical applicability. Although not specific for our method, the quality of the tumor segmentations is still too variable to be used in a fully automated radiation therapy planning pipeline. The 
proposed cRBM
tumor shape model is still fairly local, which can affect the segmentation when e.g., inner parts of the tumor have a similar appearance to healthy structures. A 
generative
model with a deeper structure, such as a variational autoencoder~\citep{kingma2013}, could potentially improve the performance on a more global scale.
 
Segmenting
one subject with the proposed method currently takes around 40 minutes. Although a manual delineation procedure is typically faster, the method can still be a useful aid in the clinical work flow, as no manual input is needed before or during the segmentation procedure. A further speed-up would be necessary to use the method for continuous segmentation during an image-guided radiation therapy session~\citep{lagendijk2014}. Since the
implementation used in this paper has mainly been focused on demonstrating the feasibility of the method rather than optimizing speed, a further speed-up would be expected with a more efficient implementation, especially with one that utilizes GPUs.

\section{Acknowledgments}
This research was supported by the NIH NCRR (P41RR14075, 1S10RR023043), NIBIB (R01EB013565) and the Lundbeck foundation (R141-2013-13117). This project has received funding from the European Union's Horizon 2020 research and innovation program under the Marie Sklodowska-Curie grant agreement No 765148. The Wellcome Centre for Human Neuroimaging is supported by core funding from the Wellcome Trust \; (203147/Z/16/Z).

\appendix
\section{Sampling from $p( \boldsymbol \theta | \mathbf{l},\mathbf{z},\mathbf{y}, \mathbf{D} )$}
\label{a:step2}
Here we describe how we sample from $p( \boldsymbol \theta | \mathbf{l},\mathbf{z},\mathbf{y}, \mathbf{D} )$ in the blocked Gibbs sampler 
used
in Section \ref{s:infer}. 

Table~\ref{t:const} specifies a number of linear constraints on 
  the Gaussian means $\{ \boldsymbol \mu_{xg}\}$ in the prior $p( \boldsymbol \theta )$,
  encoding prior knowledge about tumor appearance relative to normal brain tissue. 
  Stacking all 
  Gaussian means 
  into a single vector
  $\boldsymbol \mu
    =
    \left(
      \ldots,
      \boldsymbol \mu_{xg}^T, 
      \ldots
    \right)^T
  $  
  allows us to express these constraints in the form
  \begin{displaymath}
    \mathbf{A} \boldsymbol \mu \leq \mathbf{b}
    ,
  \end{displaymath}
  where the values in each row of $\mathbf{A}$ and $\mathbf{b}$ are chosen to 
  match
  the corresponding line in Table~\ref{t:const}.
  
Introducing the ``one-hot'' auxiliary variable
$\mathbf{t}_i = \{ t_i^{xg} \}$
to indicate which individual Gaussian component the $i^{\mathrm{th}}$ voxel is associated with
($t_i^{xg}$ has value one when the voxel belongs to the $g^{\mathrm{th}}$ component of the $x^{\mathrm{th}}$ GMM, and zero otherwise)
the target distribution 
is obtained as a marginal distribution
of \linebreak $
  p( \boldsymbol \theta, \{ \mathbf{t}_i \} | \mathbf{l},\mathbf{z},\mathbf{y}, \mathbf{D} )
$:
$
  p( \boldsymbol \theta | \mathbf{l},\mathbf{z},\mathbf{y}, \mathbf{D} ) = \sum_{\{\mathbf{t}_i\}} p( \boldsymbol \theta, \{\mathbf{t}_i\} | \mathbf{l},\mathbf{z},\mathbf{y}, \mathbf{D} )
$
.
Therefore, samples of $p( \boldsymbol \theta | \mathbf{l},\mathbf{z},\mathbf{y}, \mathbf{D} )$ 
can be obtained 
with a blocked Gibbs sampler of
$p( \boldsymbol \theta, \{\mathbf{t}_i\} | \mathbf{l},\mathbf{z},\mathbf{y}, \mathbf{D} )$,
cyclically sampling from
the following conditional distributions and 
subsequently 
discarding the samples of $\{\mathbf{t}_i\}$: 
\begin{gather}
p( \{\mathbf{t}_i\} | \boldsymbol \theta, \mathbf{l},\mathbf{z},\mathbf{y}, \mathbf{D} )
  =
  \prod_i
  p( \mathbf{t}_i | \boldsymbol \theta, x( l_i, z_i, y_i ), \mathbf{d}_i )
  \label{eq:posteriorOfT} 
  \\
\mathrm{with} \quad
  p( \mathbf{t}_i | \boldsymbol \theta, x, \mathbf{d}_i )
  = \frac{ \sum_{g=1}^{G_x}  t_i^{xg} \gamma_{xg} \mathcal{N}( \mathbf{d}_i |\boldsymbol \mu_{xg} + \mathbf{C} \boldsymbol \phi_i, \boldsymbol \Sigma_{xg} ) }{\sum_{g=1}^{G_x} \gamma_{xg} \mathcal{N}( \mathbf{d}_i |\boldsymbol \mu_{xg} + \mathbf{C} \boldsymbol \phi_i, \boldsymbol \Sigma_{xg} ) }
  \nonumber
  , \\
p(\{\boldsymbol \gamma_x\} | {\boldsymbol \theta}_{\setminus \{\boldsymbol \gamma_x\}}, \mathbf{t}, \mathbf{l},\mathbf{z},\mathbf{y}, \mathbf{D} ) 
  = \prod_x \text{Dir}\left( \boldsymbol \gamma_x | \{ \alpha_{xg} \}_{g=1}^{G_x} \right)
  ,
  \label{eq:posteriorOfGammas}
  \\
p(\{\boldsymbol \mu_{xg} \}| {\boldsymbol \theta}_{\setminus \{\boldsymbol \mu_{xg}\}}, \mathbf{t}, \mathbf{l},\mathbf{z},\mathbf{y}, \mathbf{D} ) 
\qquad \qquad \qquad \qquad \qquad
\nonumber
\\
\qquad \qquad \qquad
\propto 
\left\{
  \begin{array}{ll}
    \mathcal{N}( \boldsymbol \mu | \mathbf{m_{\mu}}, \mathbf{S_{\mu}} )
    & \mathrm{if\ } \mathbf{A} \boldsymbol \mu \leq \mathbf{b} \\
    0 & \mathrm{otherwise}
    ,
  \end{array}
\right.
\label{eq:posteriorOfMeans}
\\
  p(\{\boldsymbol \Sigma_{xg} \}| {\boldsymbol \theta}_{\setminus \{\boldsymbol \Sigma_{xg}\}}, \mathbf{t}, \mathbf{l},\mathbf{z},\mathbf{y}, \mathbf{D} ) 
  = \prod_x \prod_g \text{IW} (\boldsymbol \Sigma_{xg} | \mathbf{S}_{xg}, \upsilon_{xg})
  ,
  \label{eq:posteriorOfSigmas}
\end{gather}
and finally
\begin{align}
  p( \mathbf{C} | {\boldsymbol \theta}_{\setminus \mathbf{C}}, \mathbf{t}, \mathbf{l},\mathbf{z},\mathbf{y}, \mathbf{D} )
  = 
  \mathcal{N} ( \mathbf{c} | \mathbf{m}_c, \mathbf{S}_c)
  \quad
  \mathrm{with}
  \quad
  \mathbf{c} = \begin{pmatrix} 
                 \mathbf{c}_1 \\
                 \vdots \\
                 \mathbf{c}_N \\
               \end{pmatrix}
  .             
  \label{eq:posteriorOfBiasFieldParameters}
\end{align}
Here we have defined the following variables:
\begin{gather*}
\alpha_{xg} = \alpha_0 + N_{xg}
\quad
\mathrm{with}
\quad
N_{xg} = \sum_i t_i^{xg}
  \\
\mathbf{S_{\mu}}
  =
  \left(
    \begin{array}{ccc}
      \ddots & & \\
      & N_{xg}^{-1} \boldsymbol \Sigma_{xg} & \\
      & & \ddots
    \end{array}
  \right)
  \\ 
\mathbf{m_{\mu}}
  =
  \left(
    \begin{array}{c}
      \vdots \\
      \mathbf{m}_{xg} \\
      \vdots \\
    \end{array}
  \right)
  \quad
  \mathrm{with}
  \quad
  \mathbf{m}_{xg} = \frac{\sum_i t_i^{xg} (\mathbf{d}_i - \mathbf{C} \boldsymbol \phi_i)}{N_{xg}}
\end{gather*}
\begin{gather*}
\mathbf{S}_{xg} = \mathbf{S}_{x}^0 + \sum_i  t_i^{xg} (\mathbf{d}_i - \mathbf{C} \boldsymbol \phi_i - \boldsymbol \mu_{xg}) (\mathbf{d}_i - \mathbf{C} \boldsymbol \phi_i -  \boldsymbol \mu_{xg})^T
  \\
\upsilon_{xg} = \upsilon_{x}^0 + N_{xg}
\\ 
\mathbf{S}_c = \begin{pmatrix}
\boldsymbol \Phi^T \mathbf{W}^{11} \boldsymbol \Phi & \cdots & \boldsymbol \Phi^T \mathbf{W}^{1N} \boldsymbol \Phi \\
\vdots & \ddots & \vdots \\
\boldsymbol \Phi^T \mathbf{W}^{N1} \boldsymbol \Phi & \cdots & \boldsymbol \Phi^T \mathbf{W}^{NN} \boldsymbol \Phi \\
\end{pmatrix}^{-1}
\\
\mathrm{and}
\quad \mathbf{m}_c 
  = \mathbf{S}_c 
  \begin{pmatrix}
    \boldsymbol \Phi^T \left( \sum_{n=1}^N \mathbf{W}^{1n} \mathbf{r}^{1n} \right) \\
    \vdots \\
    \boldsymbol \Phi^T \left( \sum_{n=1}^N \mathbf{W}^{Nn} \mathbf{r}^{Nn} \right) \\
  \end{pmatrix}
  ,
\end{gather*}
\begin{gather*}
\mathrm{where}
\quad
\boldsymbol \Phi  
= 
\begin{pmatrix}
  \phi^1_1 & \cdots & \phi^1_P \\
  \vdots & \ddots & \vdots \\
  \phi^I_1 & \cdots & \phi^I_P \\
\end{pmatrix}
\quad 
\mathrm{and}
\quad
\mathbf{W}^{mn} = \text{diag} \left( w_i^{mn} \right) 
\\
\mathrm{with} \quad w_i^{mn} = \sum_x \sum_{g=1}^{G_x} w_{ixg}^{mn}, \quad w_{ixg}^{mn} = t_i^{xg} (\boldsymbol \Sigma_{xg}^{-1})_{mn}
,
\\
\quad \mathbf{r}^{mn} = (r_1^{mn}, ... , r_I^{mn} )^T,
\quad
r_i^{mn} = d_i^n - \frac{\sum_x \sum_{g=1}^{G_x} w_{ixg}^{mn} (\boldsymbol \mu_{xg} )_n}{w_i^{mn}}.
\end{gather*}

In order to sample from the truncated multivariate Gaussian distribution 
in Eq.~(\ref{eq:posteriorOfMeans}), we use the Gibbs sampling approach proposed in \citep{kotecha1999} and \citep{rodriguez2004}, 
which cycles through the conditional distributions of each component of $\boldsymbol \mu$ and samples from the corresponding truncated univariate normal distributions using inverse transform sampling. 

In our implementation, rather than repeating the Gibbs sampler steps 
described 
in Eq.~(\ref{eq:posteriorOfT}),~(\ref{eq:posteriorOfGammas}),~(\ref{eq:posteriorOfMeans}),~(\ref{eq:posteriorOfSigmas}), and~(\ref{eq:posteriorOfBiasFieldParameters})
until the Markov chain reaches equilibrium and an independent sample of $\boldsymbol \theta$ is obtained, 
we only make a single sweep before obtaining new samples of $\g$, $\h$, and $\{\labels, \z, \y \}$ in the main loop described in Algorithm \ref{a1}, effectively implementing a so-called partially collapsed Gibbs sampler~\citep{van2008partially}. 

\section{Optimizing likelihood parameters in GEM algorithm}
\label{a:step1}
Here we describe how we optimize the likelihood parameters $\boldsymbol \theta$ for a given value of the atlas node positions $\boldsymbol \eta$ 
in the simplified model of the label prior described in Section \ref{s:init}.

We use a generalized expectation-maximization (GEM) algorithm~\citep{dempster1977} that is very similar to the ones proposed in~\citep{leemput1999b} and~\citep{puonti2016}.
In short, the algorithm iteratively 
updates the various components of $\boldsymbol \theta$ to the mode of the conditional distributions 
given by 
Eq.~(\ref{eq:posteriorOfT}),~(\ref{eq:posteriorOfGammas}),~(\ref{eq:posteriorOfMeans}),~(\ref{eq:posteriorOfSigmas}), and~(\ref{eq:posteriorOfBiasFieldParameters}):
\begin{gather}
  \gamma_{xg} \gets \frac{\alpha_{xg} - 1}{\sum_{g'=1}^{G_x} (\alpha_{xg'} - 1)}, \quad \forall x,g
  \nonumber\\
  \boldsymbol \mu \gets
  \arg\max_{\boldsymbol \mu}
  \left[
  ( \boldsymbol \mu - \mathbf{m}_{\mu} )^T \mathbf{S}_{\mu}^{-1} ( \boldsymbol \mu - \mathbf{m}_{\mu} )
  \right]
  \quad
  \mathrm{s.t.}
  \quad
  \mathbf{A} \boldsymbol \mu \leq \mathbf{b}
  \label{eq:optimizeMu}
  \\
  \boldsymbol \Sigma_{xg} \gets  
   \frac{\mathbf{S}_{xg} }{\nu_{xg} + N + 1 }, \quad \forall x,g
   \nonumber\\
   \mathbf{c} \gets \mathbf{m}_c   
   \nonumber
\end{gather}
where the ``one-hot'' auxiliary variables $\{ \mathbf{t}_i \}$ 
are replaced
by their expected values: 
\begin{align}
  t_i^{xg} 
  =
  \frac{\gamma_{xg} \mathcal{N} (\mathbf{d}_i | \boldsymbol \mu_{xg} - \mathbf{C} \boldsymbol \phi_i, \boldsymbol \Sigma_{xg}) p_i (x | \boldsymbol \eta)}{\sum_{x'=1}^{X} p_i ( \mathbf{d}_i | x', \boldsymbol \theta) p_i(x'|\boldsymbol \eta)}
  , \quad \forall x,g, i
  \label{eq:posteriorOfGaussianComponents}
  .
\end{align}
Solving Eq.~(\ref{eq:optimizeMu}) is a so-called quadratic programming problem, for which an implementation is directly available in MATLAB.

\section*{References}
\biboptions{authoryear}\bibliographystyle{model2-names}
\bibliography{References}   

\end{document}